%% file: top.tex
\documentclass[10pt,twocolumn,letterpaper]{article}

\pdfoutput=1
\PassOptionsToPackage{bookmarks=false}{hyperref}

\usepackage{iccv}
\usepackage{times}
\usepackage{epsfig}
\usepackage{graphicx}
\usepackage{amsmath}
\usepackage{amssymb}

\usepackage{microtype}
\usepackage{subfigure}
\usepackage{bbm}
\usepackage{float}
\usepackage{booktabs} 
\usepackage{multirow}
\usepackage{algorithm,setspace}
\usepackage[noend]{algpseudocode}
\usepackage{array}
\usepackage{slashbox}
\usepackage[export]{adjustbox}
\usepackage{caption}

\makeatletter
\def\BState{\State\hskip-\ALG@thistlm}
\makeatother
\algdef{SE}[DOWHILE]{Do}{doWhile}{\algorithmicdo}[1]{\algorithmicwhile\ #1}%

\newcommand{\PreserveBackslash}[1]{\let\temp=\\#1\let\\=\temp}
\newcolumntype{C}[1]{>{\PreserveBackslash\centering}p{#1}}
\newcolumntype{R}[1]{>{\PreserveBackslash\raggedleft}p{#1}}
\newcolumntype{L}[1]{>{\PreserveBackslash\raggedright}p{#1}}

\usepackage[pagebackref=true,breaklinks=true,letterpaper=true,colorlinks,bookmarks=false]{hyperref}

\iccvfinalcopy 

\def\iccvPaperID{1223} 

\ificcvfinal\fi
\begin{document}

\title{\vspace{-4mm}Neural Turtle Graphics for Modeling City Road Layouts}
\author{Hang Chu$^{1,2,4}$ \enspace Daiqing Li$^{4}$ \enspace David Acuna$^{1,2,4}$ \enspace Amlan Kar$^{1,2,4}$ \enspace Maria Shugrina$^{1,2,4}$ \enspace Xinkai Wei$^{1,4}$\\
Ming-Yu Liu$^{4}$ \enspace Antonio Torralba$^{3}$ \enspace Sanja Fidler$^{1,2,4}$\\[1mm]
$^{1}$University of Toronto \qquad $^{2}$Vector Institute \qquad $^{3}$MIT \qquad $^{4}$NVIDIA\\
{\tt\scriptsize \{chuhang1122,davidj,amlan\}@cs.toronto.edu, \{daiqingl,mshugrina,xinkaiw,mingyul,sfidler\}@nvidia.com, torralba@mit.edu}
}

\makeatletter
\def\blfootnote{\gdef\@thefnmark{}\@footnotetext}
\makeatother


\newcommand{\mb}[1]{\mathbf{#1}}
\newcommand{\mrm}[1]{\mathrm{#1}}
\newcommand{\argmax}{\operatornamewithlimits{argmax}}
\definecolor{myorange}{RGB}{255,114,0}
\definecolor{mygreen}{RGB}{148,206,68}
\definecolor{myblue}{RGB}{0,150,209}

\twocolumn[{
\renewcommand\twocolumn[1][]{#1}
\maketitle
\begin{center}
\vspace{-1mm}
    \centering
\vspace{-5mm}
{\includegraphics[width=\linewidth]{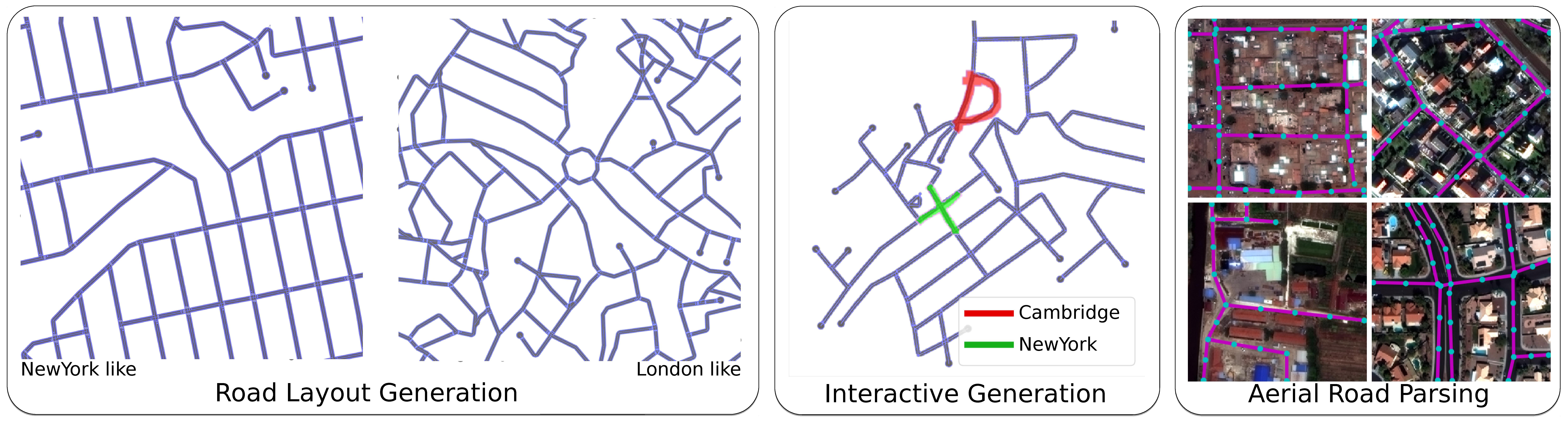}}
\vspace{-6.5mm}
    \captionof{figure}{\small
	We introduce Neural Turtle Graphics (NTG), a deep generative model for planar graphs. In the Figure, we show NTG's applications to (interactive) city road layout generation and parsing.
    	}
\end{center}
}]

\begin{abstract}
\vspace{-0.5mm}
	We propose Neural Turtle Graphics (NTG), a novel generative model for spatial graphs, and demonstrate its applications in modeling city road layouts. Specifically, we represent the road layout using a graph where nodes in the graph represent control points and edges in the graph represents road segments. NTG is a sequential generative model parameterized by a neural network. It iteratively generates a new node and an edge connecting to an existing node conditioned on the current graph. We train NTG on Open Street Map data and show that it outperforms existing approaches using a set of diverse performance metrics. Moreover, our method allows users to control styles of generated road layouts mimicking existing cities as well as to sketch  parts of the city road layout to be synthesized. In addition to synthesis, the proposed NTG finds uses in an analytical task of aerial road parsing. Experimental results show that it achieves state-of-the-art performance on the SpaceNet dataset.
\end{abstract}

\vspace{-2mm}
\input{1_intro}

\input{2_related}
\input{3_method_v2}
\input{4_experiment}
\input{4_experiment_part2}
\input{5_concolusion}

{\footnotesize
\bibliographystyle{ieee_fullname}
\bibliography{egbib}
}

\end{document}

%% file: 1_intro.tex
\vspace{-3mm}
\section{Introduction}
\vspace{-0.5mm}

	City road layout modeling is an important problem with applications in various fields. In urban planning, extensive simulation of city layouts are required for  ensuring that the final construction leads to effective traffic flow and  connectivity. Further demand comes from the gaming industry where on-the-fly generation of new environments enhances user interest and engagement. Road layout generation also plays an important role for self-driving cars, where diverse virtual city blocks are created for testing autonomous agents.

	Although the data-driven end-to-end learning paradigm has revolutionized various computer vision fields, the leading approaches~\cite{parish2001proccities} (e.g., the foundation piece in the commercially available CityEngine software) for city layout generation are still largely based on procedural modeling with hand-designed features. While these methods guarantee valid road topologies with user specified attribute inputs, the attributes are all hand-engineered and inflexible to use. For example, if one wishes to generate a synthetic city that resembles e.g. London, tedious manual tuning of the attributes is required in order to get plausible results. Moreover, these methods cannot trivially be used in aerial road parsing.
\blfootnote{Project page: {\color{magenta}{\url{https://nv-tlabs.github.io/NTG}}}}

	In this paper, we propose a novel generative model for city road layouts that learns from available map data. Our model, referred to as \emph{Neural Turtle Graphics} (NTG) is inspired by the classical turtle graphics methodology~\footnote{Turtle graphics is a technique for vector drawing, where a relative cursor (turtle) receives motion commands and leave traces on the canvas.} that progressively grows road graphs based on local statistics. We model the city road layout using a graph. A node in the graph represents a spatial control point of the road layout, while the edge represents a road segment. The proposed NTG is realized as an encoder-decoder architecture where the encoder is an RNN that encodes local incoming paths into a node and the decoder is another RNN that generates outgoing nodes and edges connecting an existing node to the newly generated nodes.  Generation is done iteratively, by pushing newly predicted nodes onto a queue, and finished once all nodes are visited. 
	Our NTG can generate road layouts by additionally conditioning on a set of attributes, thus giving control to the user in generating the content. It can also take a user specified partial sketch of the roads as input for generating a complete city road layout. Experiments with a comparison to strong baselines show that our method achieves better road layout generation performance in a diverse set of performance metrics. We further show that the proposed NTG can be used as an effective prior for aerial map parsing, particularly in cases when the imagery varies in appearance from that used in training. Fine-tuning the model jointly with CNN image feature extraction further improves results, outperforming all existing work on the Spacenet benchmark.

%% file: 2_related.tex
\section{Related Work}


\textbf{Classical Work}. A large body of literature exists on procedural modeling of streets. The seminal early work of~\cite{parish2001proccities} proposed an L-system which iteratively generates the map while adjusting parameters to conform to user guidance. This method became the foundation of the commercially available state-of-the-art CityEngine~\cite{cityengine} software.
Several approaches followed this line of work, exploiting user-created tensor fields~\cite{tensorstreet2008chen}, domain splitting~\cite{yang2013urban}, constraints stemming from the terrain~\cite{galin2011hierarchical,benes2014procedural}, and blending of retrieved examplars~\cite{aliaga2008interactive,nishida2015example,emilien2015worldbrush}. Methods that evolve a road network using constraints driven by crowd behaviour simulation have also been extensively studied~\cite{vanegas2009geobehavioral,peng2016funcspec,feng2016crowd}.


\textbf{Generative Models of Graphs.} 
Graph generation with neural networks  has only recently gained attention~\cite{you2018graphrnn,li2018learning,simonovsky2018graphvae,bojchevski2018netgan}.~\cite{you2018graphrnn} uses an RNN to generate a graph as a sequence of nodes sorted by breadth-first order, and predict edges to previous nodes as the new node is added.~\cite{simonovsky2018graphvae} uses a variational autoencoder to predict the adjacency and attribute matrices of small graphs. 
\cite{li2018learning} trains recurrent neural network that passes messages between nodes of a graph, and generates new nodes and edges using the propagated node representation. Most of these approaches only predict graph topology, while in our work we address generation of spatial graphs. Producing valid geometry and topology makes our problem particularly challenging. 
Our encoder shares similarities with node2vec~\cite{grover2016node2vec} which learns node embeddings by encoding local connectivities using random walks. Our work focuses on spatial graph generation and particularly on road layouts, thus different in scope and application.




\textbf{Graph-based Aerial Parsing}. Several work formulated road parsing as a graph prediction problem. 
The typical approach relies on CNN road segmentation followed by thinning~\cite{mattyus2017deeproadmapper}. To deal with errors in parsing, \cite{mattyus2017deeproadmapper} proposes to reason about plausible topologies on an augmented graph  as a shortest
path problem.  
In~\cite{li2018polymapper}, the authors treat local city patches as a simply connected maze which allows them to define the road as a closed polygon. Road detection then follows Polygon-RNN~\cite{CastrejonCVPR17,acuna2018efficient} which uses an RNN to predict vertices of a polygon.~\cite{homayounfar2018hierarchical} performs lane detection by predicting polylines in a top-down LIDAR view using a hierarchical RNN. Here, one RNN decides on adding new lanes, while the second RNN predicts the vertices along the lane. 
In our work, we predict the graph directly. Since our approach is local, it is able to grow large graphs which is typically harder to handle with a single RNN.
Related to our work,~\cite{marcos2018learning,DARnet19,CurveGCN19,acuna2018efficient} annotate building footprints with a graph generating neural network. However, these works are only able to handle single cycle polygons. 

Most related to our work is RoadTracer~\cite{bastani2018roadtracer}, which iteratively grows a graph based on image evidence and local geometry of the already predicted graph. At each step, RoadTracer predicts a neighboring node to the current active node. Local graph topology is encoded using a CNN that takes as input a rendering of the existing graph to avoid falling back. Our method differs in the encoder which in our case operates directly on the graph, and the decoder which outputs several outgoing nodes using an RNN which may better capture more complex road intersection topologies. Furthermore, while~\cite{bastani2018roadtracer} relied on carefully designed dynamic label creation during training to mimic their test time graph prediction, our training regime is simple and robust to test time inference. 

We also note that with some effort many of these work could be turned into generative models, however, ours is the first that showcases generative and interactive modeling of roads. Importantly, we show that NTG trained only on map data serves as an efficient prior for aerial road parsing. This cannot easily be done with existing work~\cite{bastani2018roadtracer,li2018polymapper} which all train a joint image and geometry representation. 

%% file: 3_method_v2.tex
\section{Neural Turtle Graphics}

	We formulate the city road layout generation problem as a planar graph generation problem. We first introduce the notation in Sec.~\ref{sec:notation} and then describe our NTG model in Sec.~\ref{sec:model}. Aerial parsing, implementation details, training and inference are given in Sec.~\ref{sec:parsing}-Sec.~\ref{sec:training}, respectively.  	
	
\begin{figure*}[t!]
\vspace{-1mm}
\centering
\addtolength{\tabcolsep}{3pt}
\begin{tabular}{cc}
\includegraphics[height=4.0cm,trim=00 0 0 00,clip]{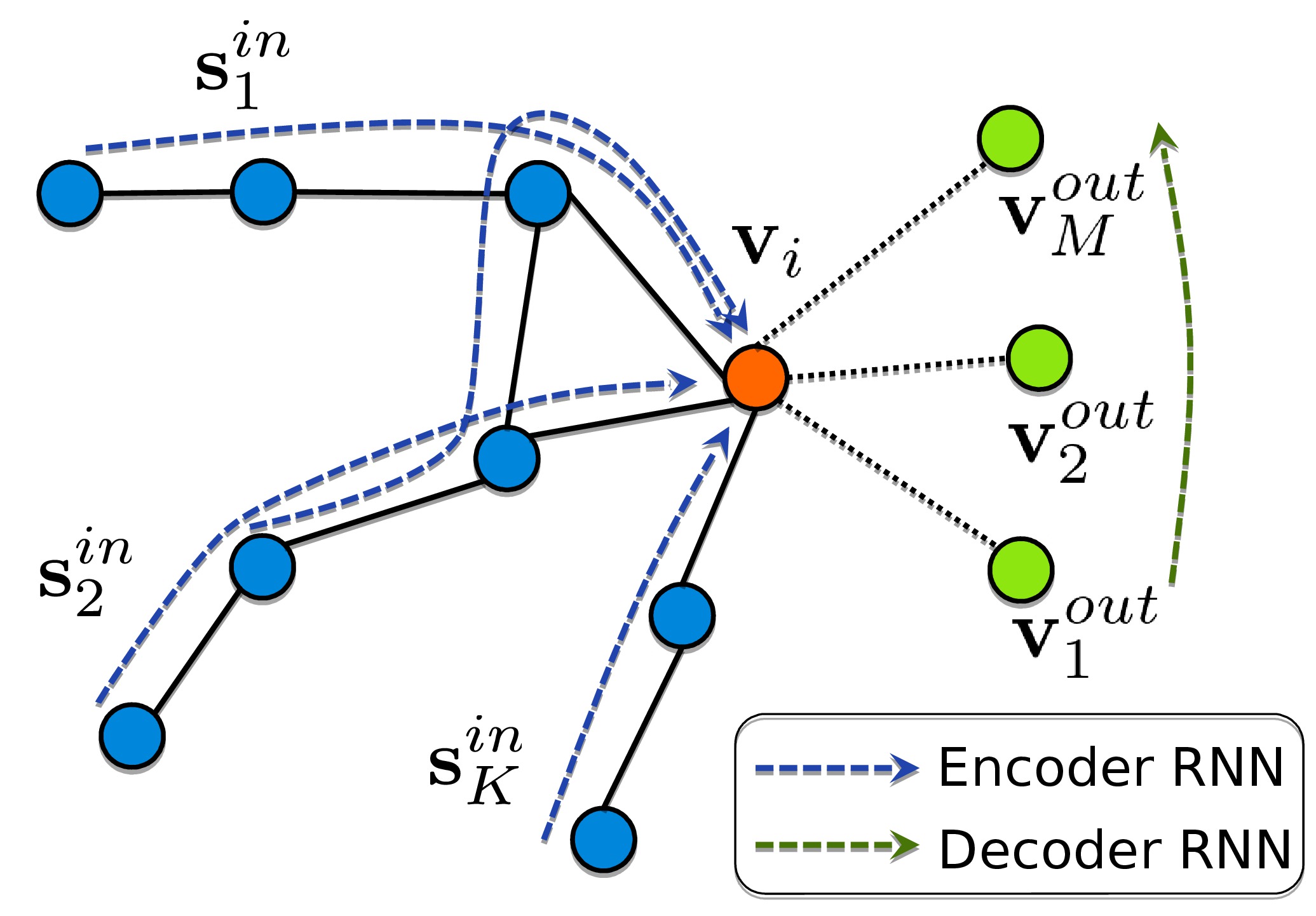} &
\includegraphics[height=4.0cm,trim=0 240 136 0,clip]{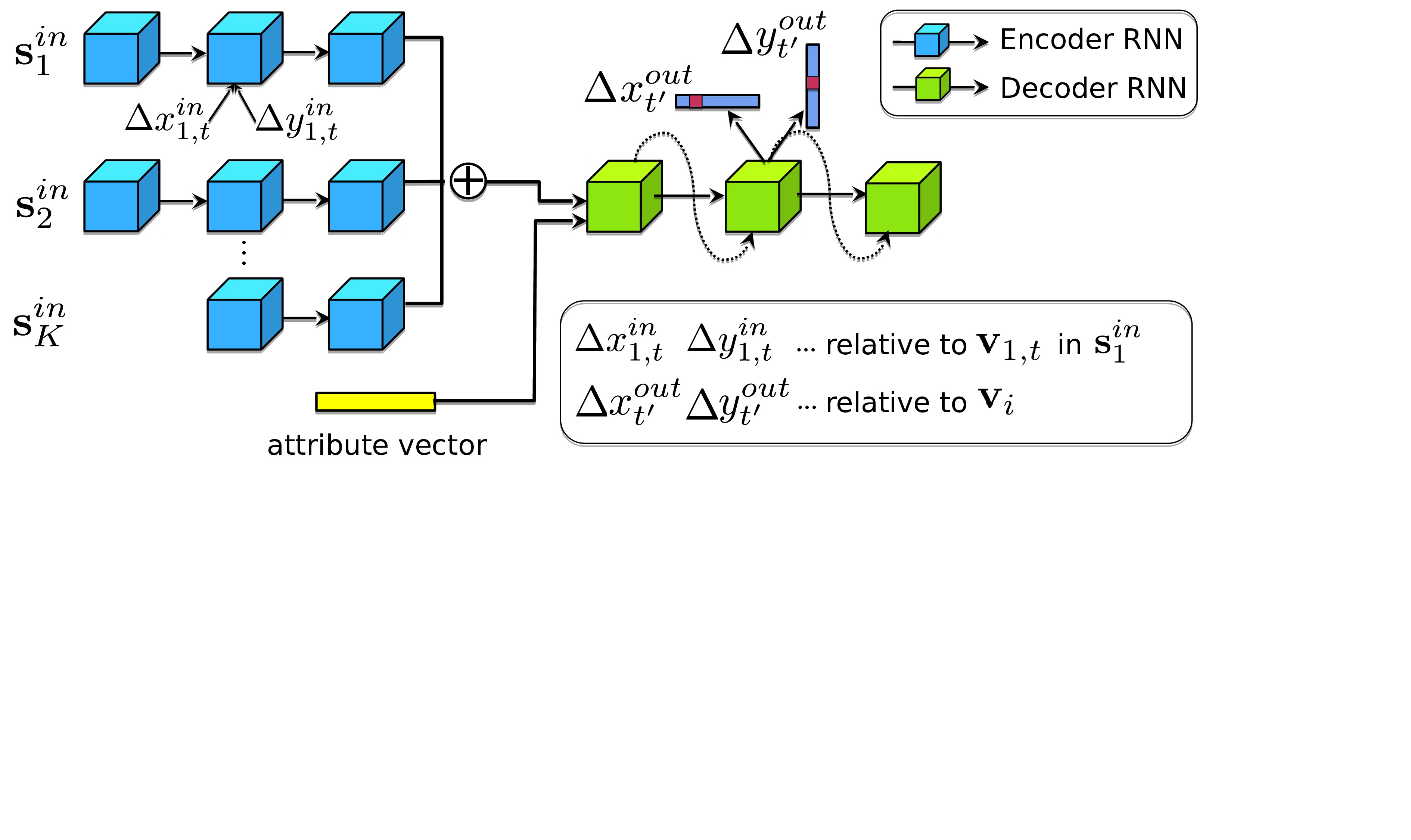}\\
(a) & (b)\\
\end{tabular}
\vspace{-3mm}
\caption{\small Illustration of the Neural Turtle Graphics (NTG) model. (a) depicts acyclic {\color{myblue}incoming paths} \{$\mb{s}^{in}$\} of an {\color{myorange} active node} $\mb{v}_i$, each of which is encoded using an RNN encoder. NTG decoder then predicts a set of {\color{mygreen} outgoing nodes} \{$\mb{v}^{out}$\}. (b) shows the NTG's neural network architecture. First, the encoder GRU consumes the motion trajectory $\Delta\mb{x}^{in}$ of each incoming path. We produce an order-invariant representation by summing up the last-state hidden vectors across all paths. Next, the decoder produces ``commands'' to advance the turtle and produces new nodes. An optional attribute vector can be further added to the decoder depending on the task.}
\label{fig:model}
\vspace{-5mm}
\end{figure*}

\subsection{Notation}
\label{sec:notation}
\noindent\textbf{Road Layout.\enspace} We represent a city road layout using an undirected graph ${G}=\{{V},{E}\}$, with nodes $V$ and edges ${E}$. A node $\mb{v}_i\in {V}$ encodes its spatial location $[x_i,y_i]^T$, while an edge $e_{\mb{v}_{i},\mb{v}_{j}}\in\{0,1\}$ denotes whether a road segment connecting nodes $\mb{v}_i$ and $\mb{v}_j$ exists. City road graphs are planar since all intersections are present in $V$. We assume it is connected, \ie there is a path in $G$ between any two nodes in $V$. The coordinates $x_i$ and $y_i$ are measured in meters, relative to the city's world location.

	\noindent\textbf{Incoming Paths.\enspace} For a node $\mb{v}_i$, we define an \textit{Acyclic Incoming Path} as an ordered sequence of unique, connected nodes which terminates at $\mb{v}_i$:
	$\mb{s}^{in}$= \{$\mb{v}_{i,1},\mb{v}_{i,2},...,\mb{v}_{i,L}, \mb{v}_{i}$\}
	where $e_{\mb{v}_{i,t},\mb{v}_{i,t+1}}=1$ for each $1\leq t < L$, and $e_{\mb{v}_{i,L},\mb{v}_{i}}=1$, with $L$ representing the length of the path. Since multiple different acyclic paths can terminate at $\mb{v}_i$, the set of these paths is denoted as $S_i^{in}:= \{\mb{s}_k^{in}\}$. 

	\noindent\textbf{Outgoing Nodes.\enspace} We define ${V}^{out}_{i} := \{\mb{v}_j: \mb{v}_j \in V \wedge e_{\mb{v}_{i},\mb{v}_{j}}=1\}$,
\ie as the set of nodes with an edge to $\mb{v}_i$.

\subsection{Graph Generation}
\label{sec:model}
	
We learn to generate graphs in an iterative manner. The graph is initialized with a root node and a few nodes connected to it, which are used to initialize a queue $Q$ of unvisited nodes. In every iteration, an unvisited node from $Q$ is picked to be expanded (called \emph{active node}). Based on its \emph{current local topology}, an encoder model generates a latent representation, which is used to generate a set of neighboring nodes using a decoder model. These generated nodes are pushed to $Q$. The node to be expanded next is picked by popping from $Q$, until it is empty.

	
By construction, an active node $\mb{v}_i$ has at least one neighbor node in the graph. NTG extracts a representation of its local topology by encoding \emph{incoming paths} $S_{i}^{in}$ (of maximum length L) and uses the representation to generate a set of \emph{outgoing nodes} $V_{i}^{out}$ (if any) with edges to $\mb{v}_i$. These paths are encoded in an order-invariant manner and the resulting latent representation is used to generate a set of outgoing nodes $V_i^{out}$. NTG performs the encoding and decoding to generate the graph as described above with an encoder-decoder neural network. Fig.~\ref{fig:model}(a) visualizes the process, while Fig.~\ref{fig:model}(b) illustrates the encoder-decoder neural architecture, which is described in detail in the following sections. 
	
	\noindent\textbf{NTG Encoder.\enspace} We encode a single incoming path $\mb{s}^{in}$ into node $\mb{v}_i$ with a zero-initialized, bidirectional GRU~\cite{chung2014empirical}. The input to the GRU while processing the $t^{th}$ node in the path is the motion vector between the nodes $\mb{v}_{i,t}^{in}\in\mb{s}^{in}$ and $\mb{v}_{i,t+1}^{in}\in\mb{s}^{in}$ in the path; \ie, $[\Delta x_{i,t}^{in},\Delta y_{i,t}^{in}]^T=[x_{i,t+1}^{in},y_{i,t+1}^{in}]^T-[x_{i,t}^{in},y_{i,t}^{in}]^T$. This offset could be encoded as a discrete or continuous value, as discussed in Sec.~\ref{sec:details}. The final latent representation $\mb{h}_\mathrm{enc}$ for all paths is computed by summing the last hidden states of each path. Optionally, we append an attribute vector $\mb{h}_\mathrm{attr}$ to the latent representation. For example, the attribute could be an embedding of a one-hot vector, encoding the city identity. This enables NTG to learn an embedding of city, enabling conditional generation. The final encoding is their concatenation $[\mb{h}_\mathrm{enc}, \mb{h}_\mathrm{attr}]$.
	
\noindent\textbf{Sampling Incoming Paths.\enspace} During training, for an active node $\mb{v}_i$ we use a subset of ${S}_{i}^{in}$ by sampling $K$ random walks (without repetition) starting from $\mb{v}_i$, such that each random walk visits at most $L$ different nodes. We find this random sampling to lead to a more robust model as it learns to generate from incomplete and diverse input representations.
Optionally, we can also feed disconnected adjacent nodes as additional input. We found this to perform similarly in the task of road modeling due to high connectivity in data.

	\noindent\textbf{Decoder.\enspace} We decode the outgoing nodes ${V}^{out}_i$ with a decoder GRU. The recurrent structure of the decoder enables capturing local dependencies between roads such as orthogonality at road intersections. We independently predict $\Delta x_{t'}^{out}$ and $\Delta y_{t'}^{out}$ for an outgoing node $\mb{v}_{t'}^{out}$, indicating a new node's relative location w.r.t. $\mb{v}_i$. Additionally, we predict a binary variable which indicates whether another node should be generated. At generation time we check overlap between the new node and existing graph with a 5$m$ threshold to produce loops. Optionally, we predict the edge type between $(\mb{v}_i, \mb{v}_{t'}^{out})$, \ie minor or major road, using a categorical variable. The hidden state $\mb{h}_{t'}$ of the decoder is updated as:
	\begin{equation}
	\mb{h}_{t'+1}=\mathrm{GRU}(\mb{h}_{t'}\mid \mb{h}_\mathrm{enc}, \mb{h}_\mathrm{attr}, \Delta\mb{x}_{t'}^{out})
	\end{equation}

\subsection{Aerial Road Parsing}
\label{sec:parsing}

\noindent\textbf{Parsing with Map Prior.\enspace} The dominant approaches to parse roads from aerial imagery have trained CNNs to produce a probability (likelihood) map, followed by thresholding and thinning. This approach typically results in maps with holes or false positive road segments, and heuristics are applied to postprocess the topology. We view a NTG model trained to generate city graphs (\ie \emph{not trained} for road parsing) as a prior, and use it to postprocess the likelihood coming from the CNN. 
Starting from the most confident intersection node as the root node, we push all its neighbors into the queue of unvisited nodes, and then use NTG to expand the graph. At each decoding step, we multiply the likelihood from CNN with the prior produced by NTG, and sample output nodes from this joint distribution. The end of sequence is simply determined by checking whether the maximum probability of a new node falls below a threshold (0.05 in our paper).
 
 
\noindent\textbf{Image-based NTG.\enspace} We also explore \emph{explicitly training} NTG for aerial parsing. We condition on image features by including predictions from a CNN trained to parse aerial images in the attribute vector $\mb{h}_{attr}$. In practice, we initialize the graph obtained by thresholding and thinning the ouputs of the CNN, and use the trained image-based NTG on top.


\subsection{Implementation Details}
\label{sec:details}
We exploit the same NTG model in both tasks of city generation and road detection. Depending on the task, we empirically find that the best parameterization strategy varies. For city generation, we use discrete $\Delta x$, $\Delta y$ with resolution of 1\textit{m} for both encoder and decoder, where $x$ points to east and $y$ points to north. Here, $\Delta x$ and $\Delta y$ are limited to $[-100:100]$, indicating that the largest offset in either direction is 100\textit{m}. The discrete $\Delta x$ and $\Delta y$ values are given associated embeddings (resembling words in language models), which are concatenated to generate the input to the encoder at every step. For road detection, we use continuous polar coordinates in the encoder, where the axis is rotated to align with the edge from the previous to the current node. This forms rotation invariant motion trajectories that help detecting roads with arbitrary orientation. The decoder always uses a discrete representation. We encode and decode the coordinates $x$ and $y$ independently. We find that this yields similar results compared to joint prediction, while significantly saving training memory and model capacity. 500 hidden units are used in encoder and decoder GRUs.

\subsection{Learning \& Inference}
\label{sec:training}

\noindent\textbf{Inference.\enspace} At each inference step, we pop a node from the queue $Q$, encode its existing incoming paths, and generate a set of new nodes. For each new node, we check if it is in the close proximity of an existing node in the graph. If the distance to an existing node is below a threshold $\epsilon$ (5\textit{m} in our paper), we do not add the new node to the queue. Instead, an edge is included to connect the current node to the existing node. This enables the generation of cycles in the graph. We also find the maximum node degree, maximum node density, and minimum angle between two edges in the training set, and ensure our generation does not exceed these limits. We refer to supplemental material for their effects.

	\noindent\textbf{Learning.\enspace} At training time, $K$ incoming paths for each $\mb{v}_i$ are sampled, and we learn to predict \emph{all} of its neighboring nodes. 
We enforce an order in decoding the nodes, where we sort nodes counter-clockwise to form a sequence. The ordering saves having to solve an assignment problem to compute the loss function. Our model is trained using ground truth map data with teacher-forcing~\cite{williams1989learning}, using a cross entropy loss for each of the output nodes.  The networks are optimized using Adam~\cite{kingma2014adam} with a learning rate of 1e-3, weight decay of 1e-4, and gradient clipping of 1.0.


 

%% file: 4_experiment.tex
\section{Experiments}

\begin{table}[t!]
\centering
\setlength{\tabcolsep}{2pt}
\begin{tabular}{c||cccccc}
	\hline
	~ & \small{Country} & \small{City} & \small{Node} & \small{Edge} & \small{Area} & \small{Length}\\
	\hline
	\hline
	RoadNet & \small{13} & \small{17} & \small{233.6\textit{k}} & \small{262.1\textit{k}} & \small{170.0\textit{km}$^2$} & \small{7410.7\textit{km}}\\
	\hline
	SpaceNet & \small{4} & \small{4} & \small{115.8\textit{k}} & \small{106.9\textit{k}} & \small{122.3\textit{km}$^2$} & \small{2058.4\textit{km}}\\
	\hline
\end{tabular}
\vspace{-3mm}
\caption{\small Dataset statistics of RoadNet and SpaceNet~\cite{spacenet}.}
\label{tab:stat}
\vspace{-6mm}
\end{table}

\begin{table*}[t!]
\centering
\addtolength{\tabcolsep}{-2pt} 
\begin{tabular}{l||L{1.0cm}L{1.0cm}L{1.0cm}L{1.0cm}|L{0.8cm}||L{1.0cm}L{1.0cm}L{1.0cm}L{1.0cm}|L{0.8cm}||l}
	\hline
	\multirow{3}{*}{\backslashbox{\small{Method}}{\small{Metric}}} & \multicolumn{5}{c||}{\textit{\small{Perceptual}}} & \multicolumn{5}{c||}{\textit{\small{Urban Planning}}} & \multirow{3}{*}{\textit{\small{Diversity}}}\\
	\cline{2-11}
	& \tt{\small{mp1}} & \tt{\small{mp2}} & \tt{\small{pa}} & \tt{\small{fc}} & \multirow{2}{*}{\tt{\small{rate}}} & \tt{\small{densi.}} & \tt{\small{conne.}} & \tt{\small{reach}} & \tt{\small{conve.}} & \multirow{2}{*}{\tt{\small{rate}}} &\\
	& \small{$10^{-1}$} & \small{$10^{0}$} & \small{$10^{-1}$} & \small{$10^{1}$} & & \small{$10^{1}$} & \small{$10^{-2}$} & \small{$10^{5}$} & \small{$10^{-3}$} & &\\
	\hline
	\hline
	GraphRNN-2D~\cite{you2018graphrnn,bastani2018roadtracer} & 7.12 & 6.35 & 8.45 & 16.15 & 25.0 & 51.58 & 4.61  & 45.11 & 6.72  & 43.7 & 44.26 \\
	PGGAN~\cite{karras2017progressive} & 1.98 & 2.15 & 5.34 & 10.51 & 63.2 & 45.77 & 19.48 & 4.33  & 2.94  & 58.9 & \underline{5.95} \\
	CityEngine-5k~\cite{cityengine}  & 2.74 & 2.71 & 8.34 & 14.78 & 47.1 & 13.59 & 21.66 & 7.61  & 16.66 & 51.7 & 45.86 \\
	CityEngine-10k~\cite{cityengine} & 2.55 & 2.56 & 8.23 & 14.17 & 48.9 & 12.43 & 21.79 & 7.05  & 16.82 & 52.1 & 46.00 \\
	\hline
	NTG-vanilla    & 2.63 & 2.33 & 4.05 & 9.17  & 66.0 & 8.69  & \textbf{1.87} & 8.99  & 3.06 & 86.5 & 41.27 \\
	NTG-enhance & \textbf{1.52} & \textbf{1.34} & \textbf{2.83} & \textbf{6.76} & \textbf{77.3} & \textbf{3.76} & 1.97 & \textbf{4.13} & \textbf{1.86} & \textbf{92.4} & 42.09 \\
	\hline
\end{tabular}
\vspace{-3mm}
\caption{\small Perceptual domain-adapted FIDs (\{maxpool1,maxpool2,pre-aux,fc\}, lower is better), Urban Planning feature differences (\{density,connectivity,reach,convenience\}, lower is better), and Diversity evaluation of city generation. Ratings (higher is better) are computed by averaging with scales \{10,10,10,20\} for perceptual and \{60,30,50,20\} for urban planning. \underline{Extremely low} Diversity indicates incapability of creating new cities.}
\label{tab:cityresult}
\vspace{-3mm}
\end{table*}

\begin{figure*}[t!]
\vspace{-1mm}
\begin{center}
\setlength{\tabcolsep}{4pt}
\begin{tabular*}{0.9\textwidth}{|c||c|c|c|c|c|}
\hline
~ & \scriptsize{GraphRNN-2D~\cite{you2018graphrnn,bastani2018roadtracer}} & \scriptsize{PGGAN~\cite{karras2017progressive}} & \scriptsize{CityEngine~\cite{cityengine}} & \scriptsize{NTG} & \scriptsize{GT}\\
\hline
\midrule[1pt]
\rotatebox{90}{NewYork} &
{\includegraphics[width=0.154\linewidth]{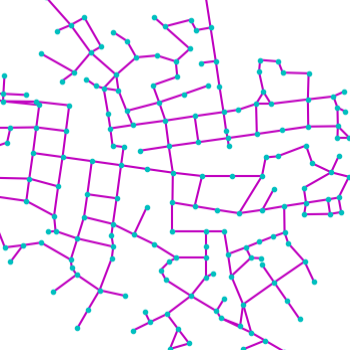}}&
{\includegraphics[width=0.154\linewidth]{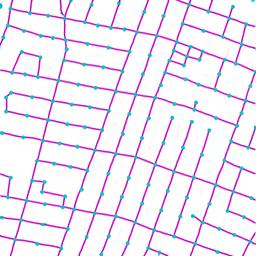}}&
{\includegraphics[width=0.155\linewidth]{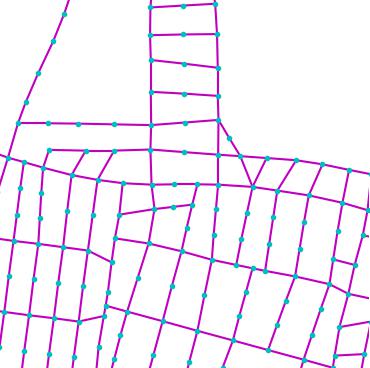}}&
{\includegraphics[width=0.155\linewidth]{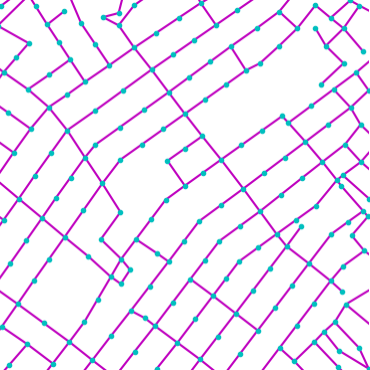}}&
{\includegraphics[width=0.155\linewidth]{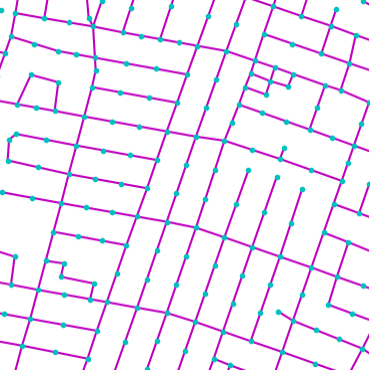}}\\
\midrule[1pt]
\rotatebox{90}{MexicoCity} &
{\includegraphics[width=0.154\linewidth]{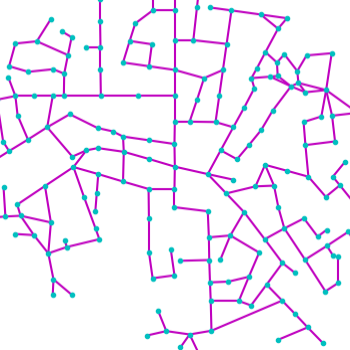}}&
{\includegraphics[width=0.154\linewidth]{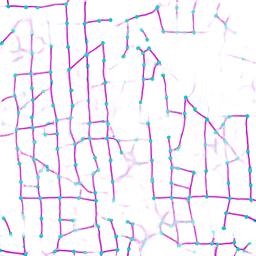}}&
{\includegraphics[width=0.155\linewidth]{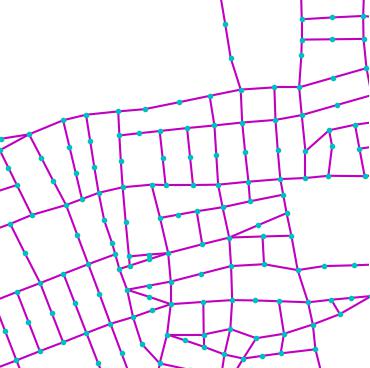}}&
{\includegraphics[width=0.155\linewidth]{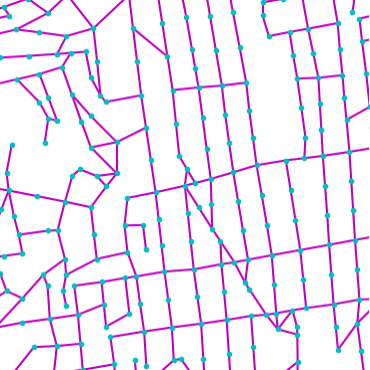}}&
{\includegraphics[width=0.155\linewidth]{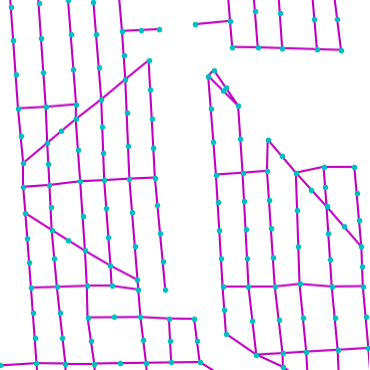}}\\
\midrule[1pt]
\rotatebox{90}{Istanbul} &
{\includegraphics[width=0.154\linewidth]{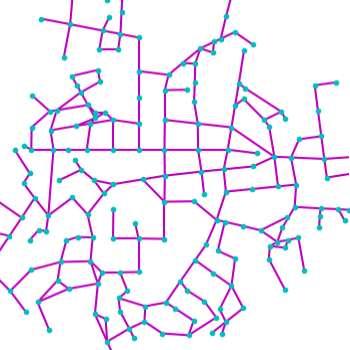}}&
{\includegraphics[width=0.154\linewidth]{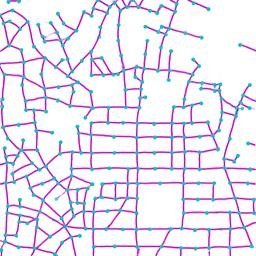}}&
{\includegraphics[width=0.155\linewidth]{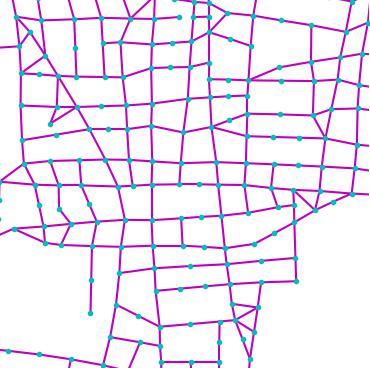}}&
{\includegraphics[width=0.155\linewidth]{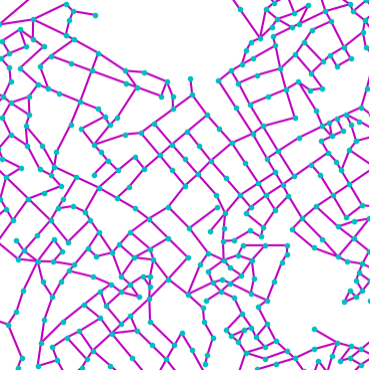}}&
{\includegraphics[width=0.155\linewidth]{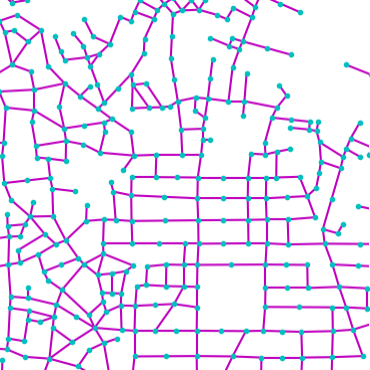}}\\
\midrule[1pt]
\end{tabular*}
\end{center}
\vspace{-7mm}
\caption{\small Qualitative examples of city road layout generation. GraphRNN-2D generates unnatural structures and fails to capture city style. PGGAN is unable to create new cities by either severely overfitting, or producing artifacts. CityEngine produces less style richness due to its fixed rule-based synthesis algorithm. NTG is able to both capture the city style and creating new cities.}
\label{fig:cityexample}
\vspace{-6mm}
\end{figure*}

We demonstrate NTG on three tasks: city road layout generation, satellite road parsing, and environment simulation. 

\subsection{City Road Layout Generation}
\noindent\textbf{RoadNet Dataset.\enspace} We collected a real-world road dataset from OpenStreetMap (OSM) to facilitate this task. In particular, we selected 17 unique cities across continents and gathered all road markers. OSM, being crowd-sourced, often has incomplete markers in underpopulated areas. To alleviate this, we manually select the most densely annotated 10\textit{km}$^2$ region within each city. These constitute our final RoadNet dataset. Table~\ref{tab:stat} shows the statistics.
\vspace{-4mm}
\subsubsection{Metrics}
\vspace{-2mm}
The goals of road layout generation are to create road networks that are: {\bf a)} Perceptually plausible and preserve a meaningful city style, and {\bf b)} Diverse. We use three broad categories of automatic metrics to evaluate city generation: 

{\bf Perceptual}: For every node, we render the graph in a 300\textit{m} neighborhood centered around it on a canvas. With their perceptual features extracted from an  InceptionV3 network~\cite{szegedy2016rethinking}, we compute the Fr\'echet Inception Distance (FID)~\cite{heusel2017gans} between the synthesized roads and ground truth maps for each city. To ensure a meaningful FID, we adapt InceptionV3, which has originally been trained on natural images, to road drawings, by finetuning it to predict city ids on our dataset. This yields a 90.27\% accuracy, indicating effective capture of style across cities in the network. 

{\bf Urban Planning}~\cite{alhalawani2014makes}: We measure four common urban planning features reflecting city style: {\bf 1.} Node \emph{density} within a neighborhood of 100\textit{m}, 200\textit{m}, and 300\textit{m}. {\bf 2.} \emph{Connectivity} as reflected by the degrees of nodes. {\bf 3.} \emph{Reach} as the total length of accessible roads within a distance of 100\textit{m}, 200\textit{m}, and 300\textit{m}. {\bf 4.)} Transportation \emph{convenience} as the ratio of the euclidean distance over the Dijkstra shortest path for node pairs that are more than 500\textit{m} away. We also compute the Fr\'echet distance between normal distributions computed from the concatenation of the Urban Planning features of real and generated maps.

{\bf Diversity metric}: We measure the ability to create novel cities by computing the overlap between a real and generated city as the percentage of road in one graph falling outside the 10\textit{m} vicinity of the road in the other graph, and vice versa. We compare this Chamfer-like distance against all ground truth maps and report the average lowest value.
\vspace{-5mm}
\subsubsection{Results}
\vspace{-2mm}
We compare the following methods:

\noindent$\bullet$ \textbf{GraphRNN-2D~\cite{you2018graphrnn,bastani2018roadtracer}}: We enhance the GraphRNN model by introducing extra branches to encode/decode node coordinates and city id. We add a CNN that takes into account local rendering of existing graph as in~\cite{bastani2018roadtracer} and add checks to avoid invalid edge crossing during inference. 

\noindent$\bullet$ \textbf{PGGAN~\cite{karras2017progressive}}: We train to generate images of road layouts at a resolution of 256$\times$256. We use our trained InceptionV3 network to classify samples into cities to compute city-wise metrics.
For computing the graph-related metrics we convert images to graphs by thresholding and thinning.

\noindent$\bullet$ \textbf{CityEngine~\cite{cityengine}}: CityEngine is a state-of-the-art software for synthesizing cities based on an optimized, complex rule-based L-system~\cite{parish2001proccities}. By default, it only offers limited templates and is incapable of generating new cities. To enhance CityEngine, we use its provided control interface and exhaustively search over its attribute space by enumerating combinations of important control parameters such as angles, bending specifications, and crossing ratios. We then predict city probabilities using the InceptionV3 network, and select the highest ranking 10\textit{km}$^{2}$ as the result for each city. 

\begin{figure}[t!]
\centering
\includegraphics[width=\linewidth]{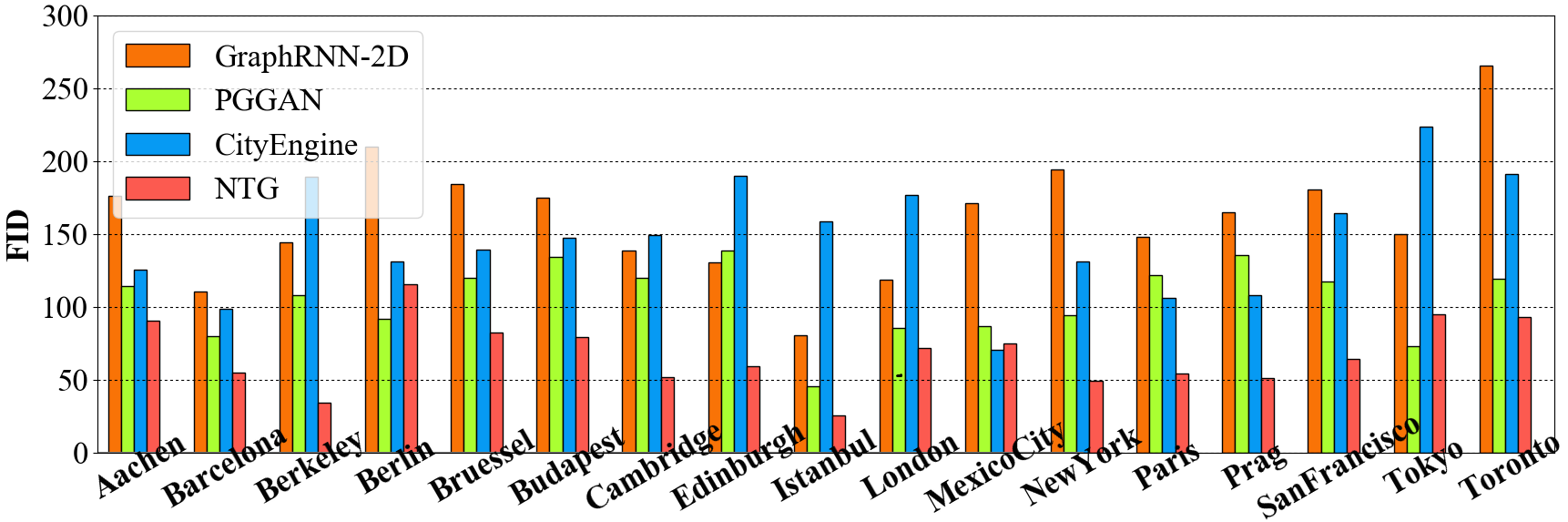}
\vspace{-7mm}
\caption{\small City-wise FID (fc) of different methods.}
\label{fig:percity}
\vspace{-7mm}
\end{figure}
\vspace{1mm}
\noindent$\bullet$ \textbf{NTG}: NTG begins with a root node with its edges. We evaluate NTG with a random root (NTG-vanilla), as well as with a pre-stored high connectivity root (NTG-enhance).

Tab.~\ref{tab:cityresult} and Fig.~\ref{fig:cityexample} show quantitative and qualitative results, respectively. Quantitatively, NTG outperforms baselines across all metrics. GraphRNN-2D fails to capture various city styles and frequently produces unnatural structures.
This is due to its sequential generative nature that depends on Breadth First Search. The RNN that encodes the full history fails to capture coherent structures since consecutive nodes may not be spatially close due to BFS. PGGAN~\cite{karras2017progressive} produces sharp images with occasional artifacts that are difficult to convert into meaningful graphs. Samples from PGGAN are severely overfit as reflected by the Diversity metric, indicating its inability to create new cities. Moreover, PGGAN also suffers from mode-collapse and memorizes a portion of data. This imbalance of style distribution leads to worse perceptual FIDs. With our enhancement (exhaustive search), CityEngine~\cite{cityengine} is able to capture certain cities' style elements: especially the node density. However, it has less topological variance and style richness due to its fixed set of rules. Expanding its search from 5000\textit{km}$^{2}$ (CityEngine-5k) to 10000\textit{km}$^{2}$ (CityEngine-10k) of generated maps does not lead to significant improvements, while requiring double the amount of computation. NTG is able to create new cities, while better capturing style in most cities as shown in Fig.~\ref{fig:percity}. 

\begin{figure}[h!]
\vspace{-1mm}
\begin{center}
\begin{tabular}{c|c}
\hline
\small{GT} & \small{NTG}\\
\hline
\midrule[1pt]
{\includegraphics[width=0.35\linewidth]{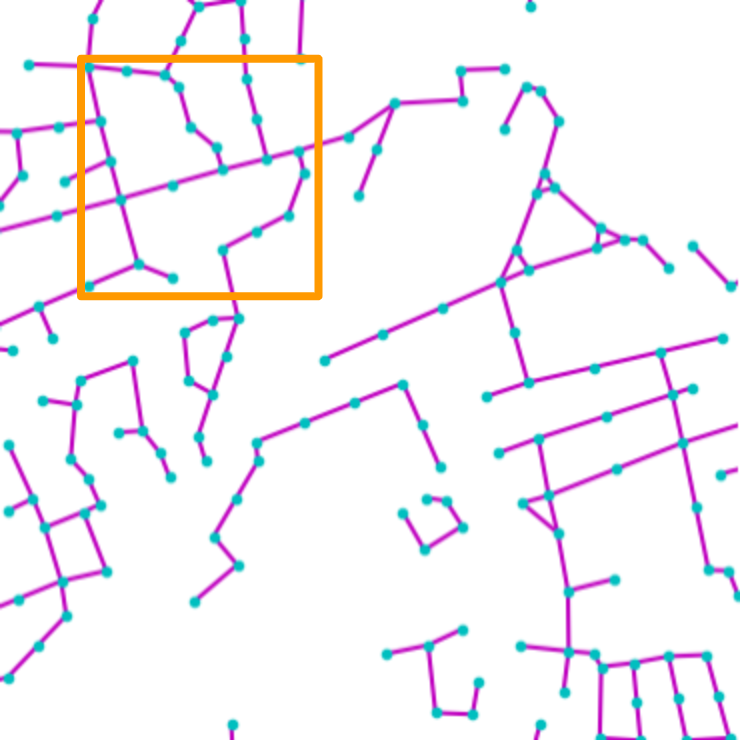}}&
{\includegraphics[width=0.35\linewidth]{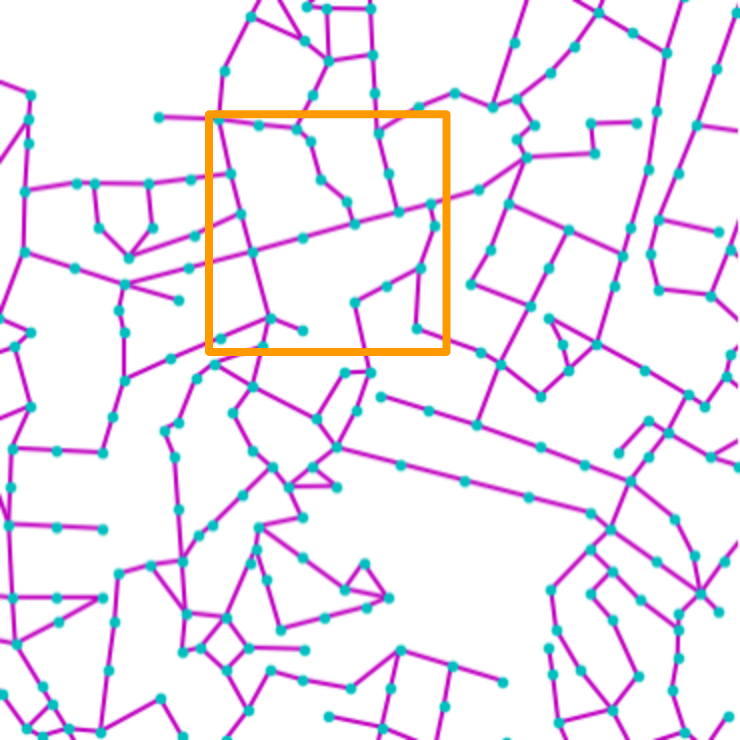}}\\
\midrule[1pt]
\end{tabular}
\end{center}
\vspace{-7mm}
\caption{\small NTG creates new city road layouts in a combinatorial manner. Local patterns as shown by orange boxes are remembered, and then intertwined to create novel structures.}
\label{fig:cityanalyse}
\vspace{-3mm}
\end{figure}

\begin{figure}[h!]
\centering
\setlength{\tabcolsep}{0pt} 
\begin{tabular}{ccc}
\includegraphics[height=3.0cm]{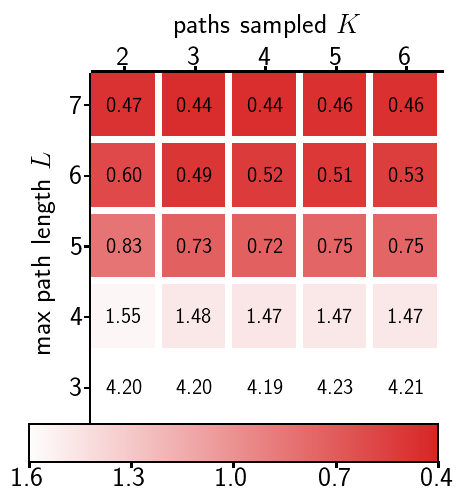} &
\includegraphics[height=3.0cm]{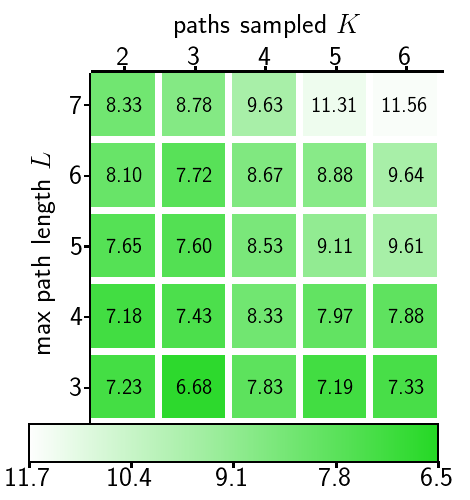} &
\includegraphics[height=3.0cm]{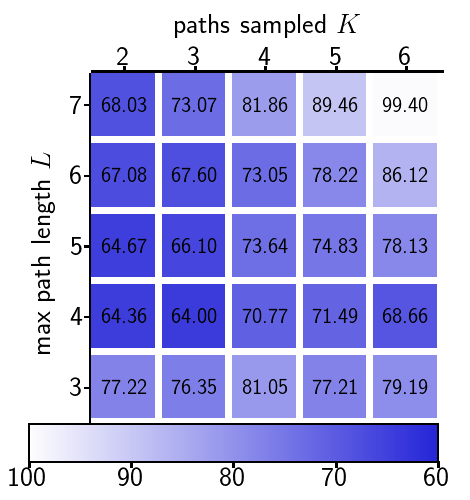}\\
\end{tabular}
\vspace{-4mm}
\caption{\small Effect of sampled paths $K$ and maximum path length $L$ on reconstruction quality in meters (red), inference time in seconds per \textit{km}$^{2}$ (green), and FID-fc (blue).}
\label{fig:cityhyper}
\vspace{-6mm}
\end{figure}

Fig.~\ref{fig:cityanalyse} digs into NTG's generated maps, showing that NTG learns to remember local road patterns. As the graph grows, different local topologies are organically intertwined to produce new cities. Fig.~\ref{fig:cityhyper} shows the effect of two important hyper-parameters: maximum number of paths $K$ and maximum incoming path length $L$. Results show that reconstruction quality is determined by $L$, while $K$ and $L$ both affect inference time. Training with longer and more paths does not necessarily improve perceptual quality, since generation starts from a single root node without long paths.

\begin{figure}[t!]
\vspace{-2mm}
\begin{center}
\begin{tabular}{c|c}
\hline
\small{Brussels} & \small{Toronto}\\
\hline
\midrule[1pt]
{\includegraphics[width=0.35\linewidth]{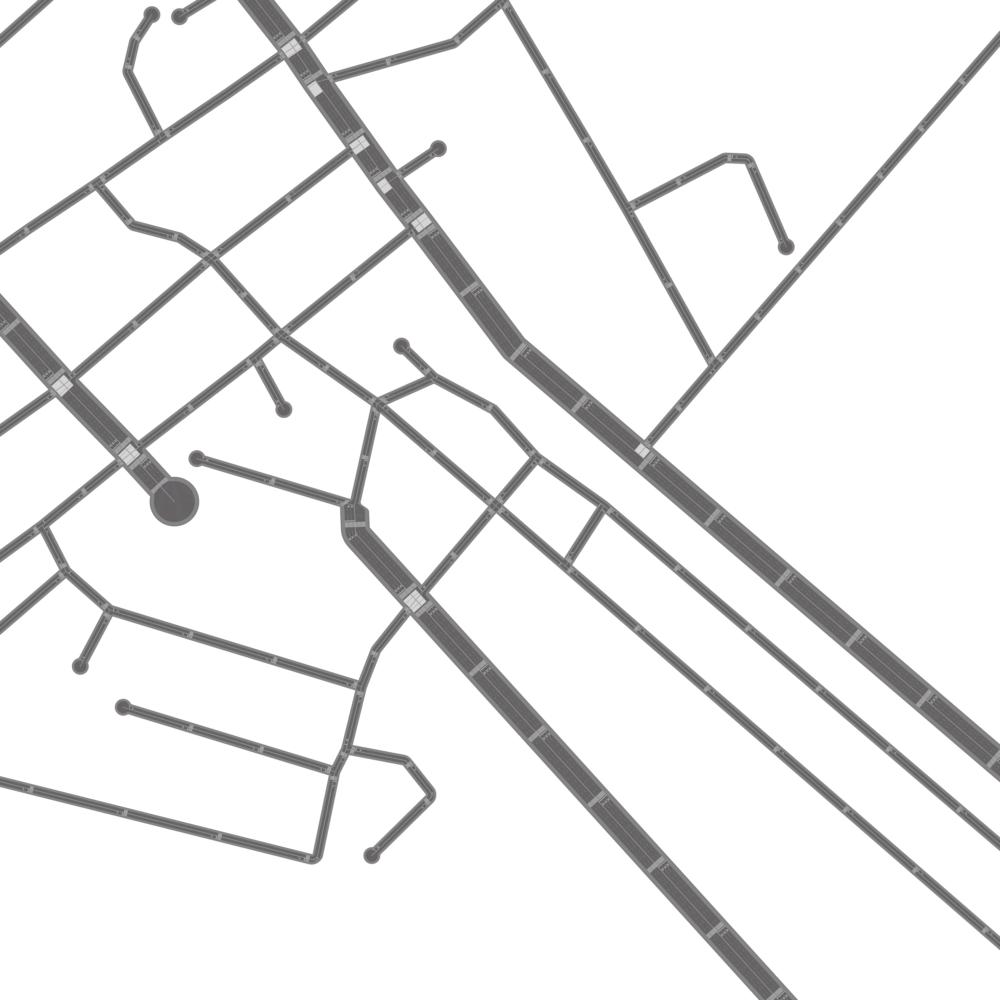}}&
{\includegraphics[width=0.35\linewidth]{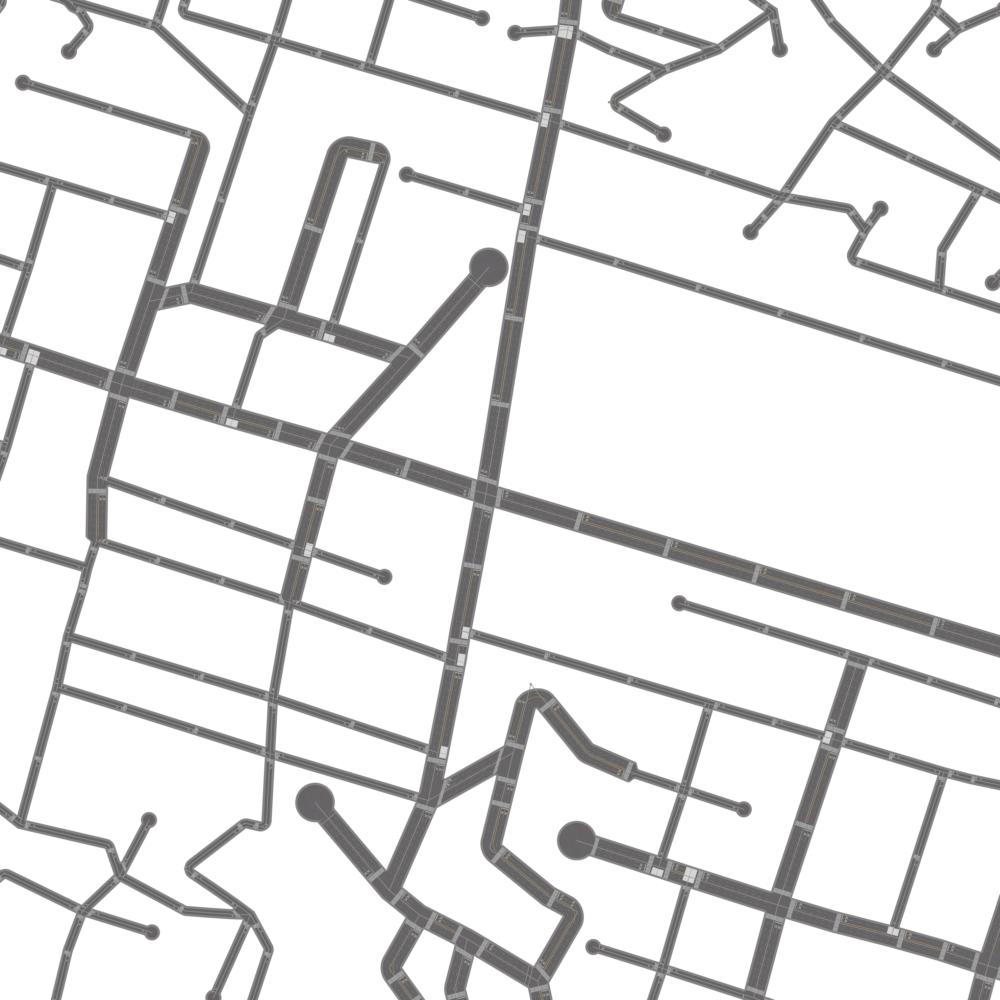}}\\
\midrule[1pt]
\end{tabular}
\end{center}
\vspace{-7mm}
\caption{\small NTG can be easily extended to generate road type.}
\label{fig:twotype}
\vspace{-4mm}
\end{figure}

We further demonstrate our approach by having NTG predict two types of roads, \ie major and minor roads. Results are shown in Fig.~\ref{fig:twotype}, showing that NTG easily generalizes to a more complex modeling problem. 

\begin{figure*}[h!]
\vspace{-1mm}
\begin{center}
\setlength{\tabcolsep}{2pt}
\begin{tabular}{|c||c|c|c|c|c||c|}
\hline
\small{user input} & \small{step 0} & \small{step 20} & \small{step 40} & \small{step 60} & \small{step 80} & \small{final}\\
\hline
\midrule[1pt]
{\includegraphics[width=0.115\linewidth]{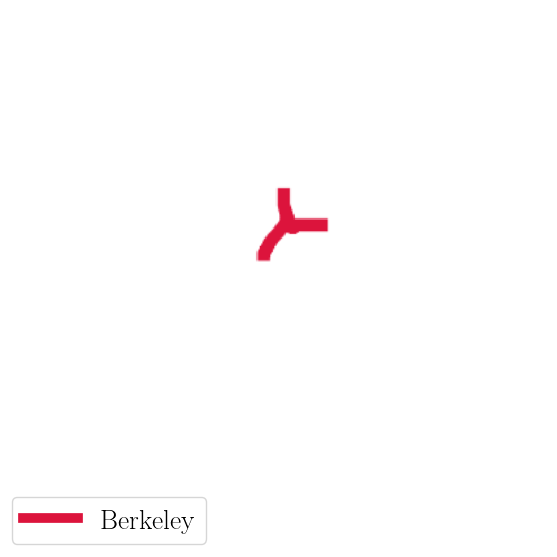}}&
{\includegraphics[width=0.115\linewidth]{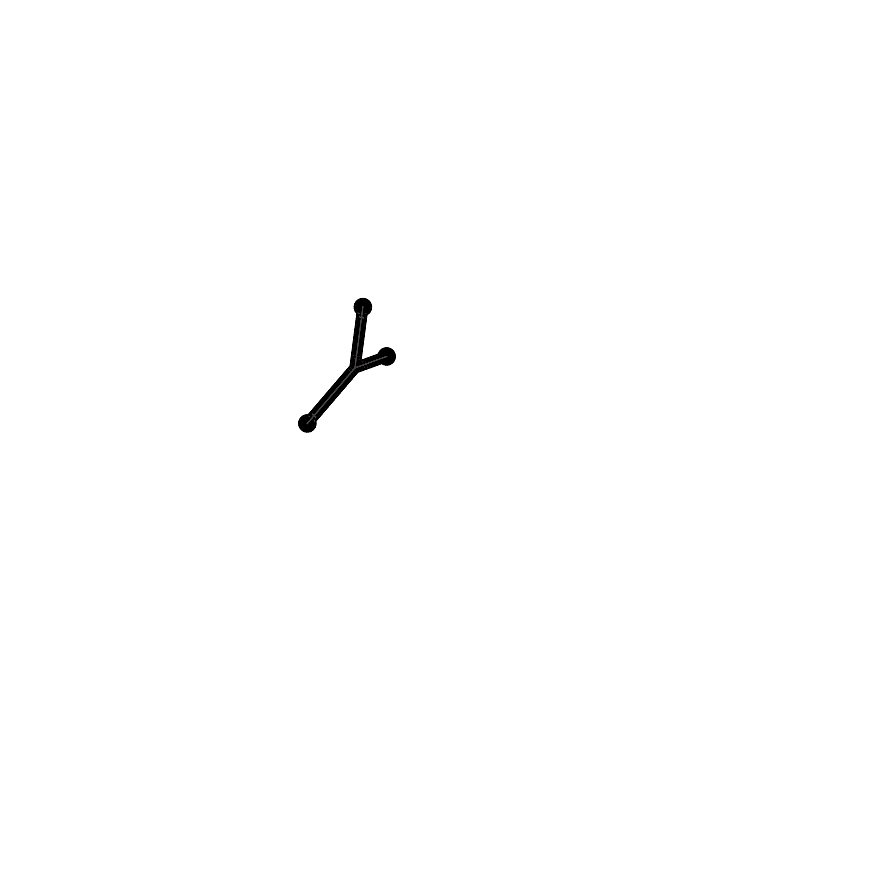}}&
{\includegraphics[width=0.115\linewidth]{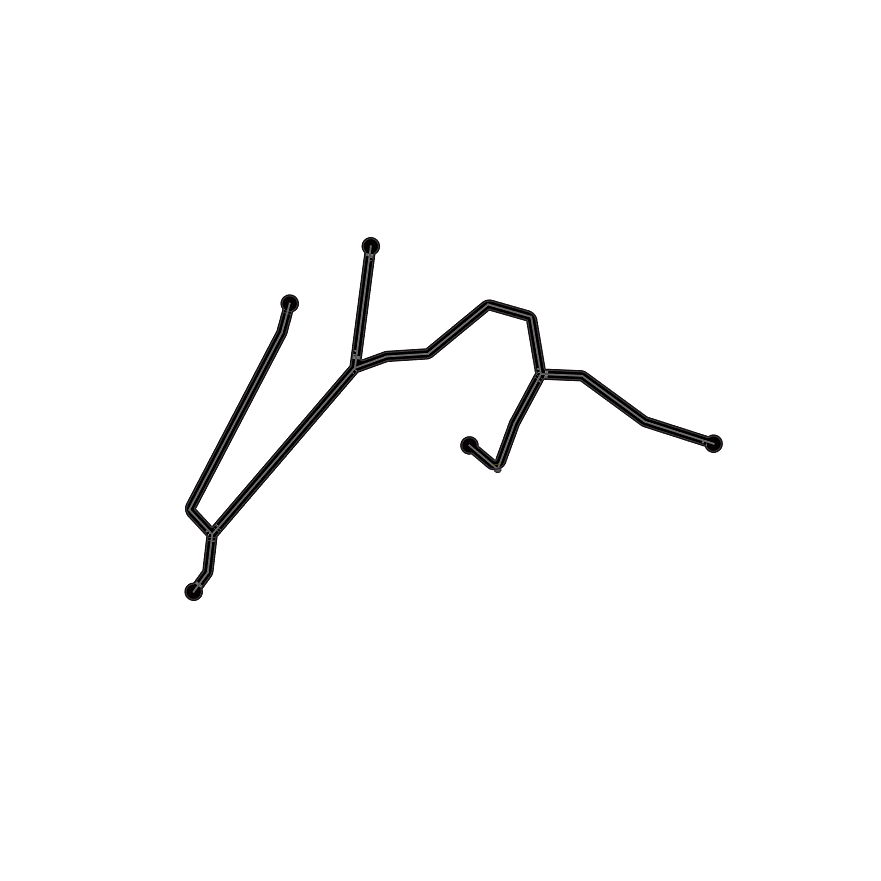}}&
{\includegraphics[width=0.115\linewidth]{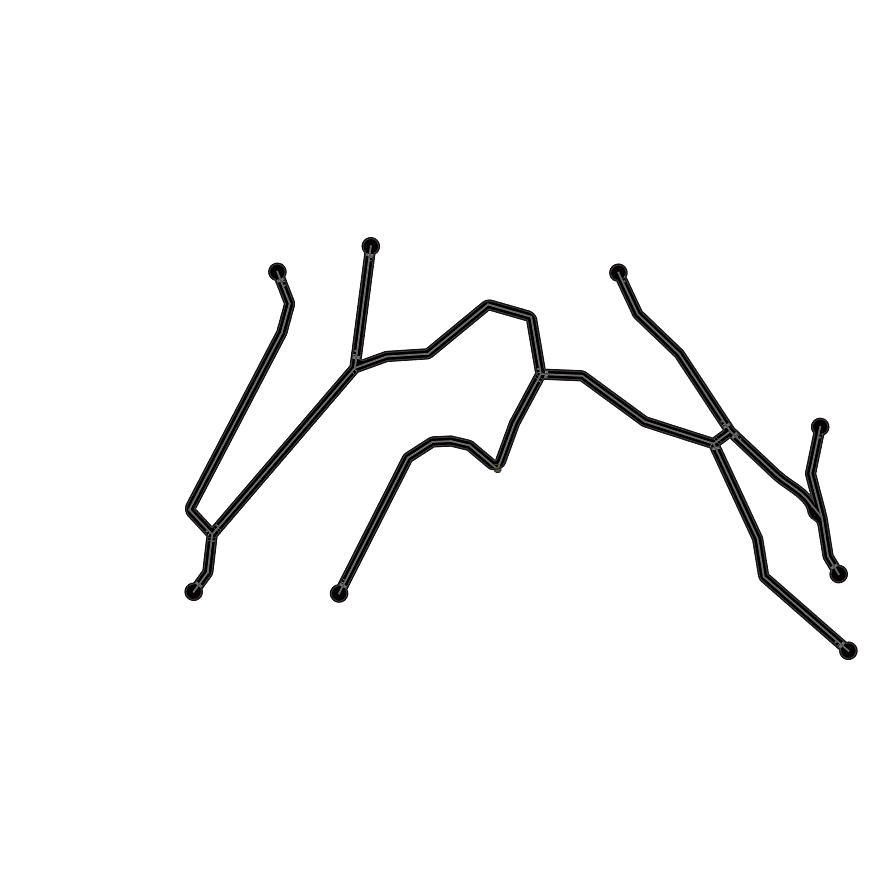}}&
{\includegraphics[width=0.115\linewidth]{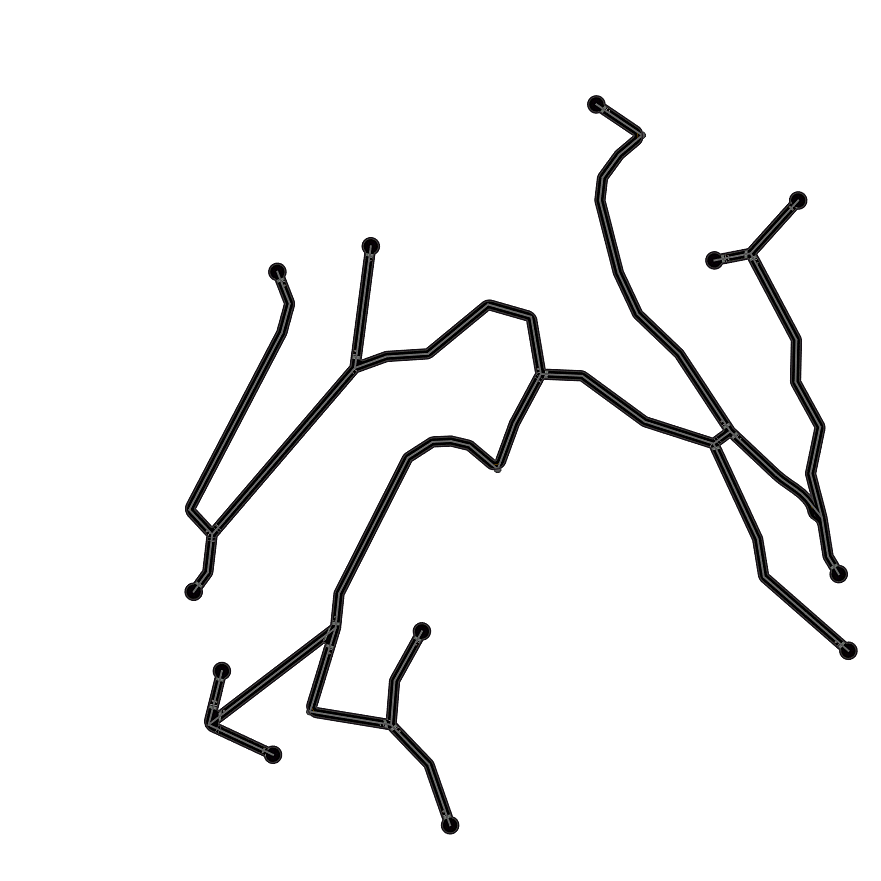}}&
{\includegraphics[width=0.115\linewidth]{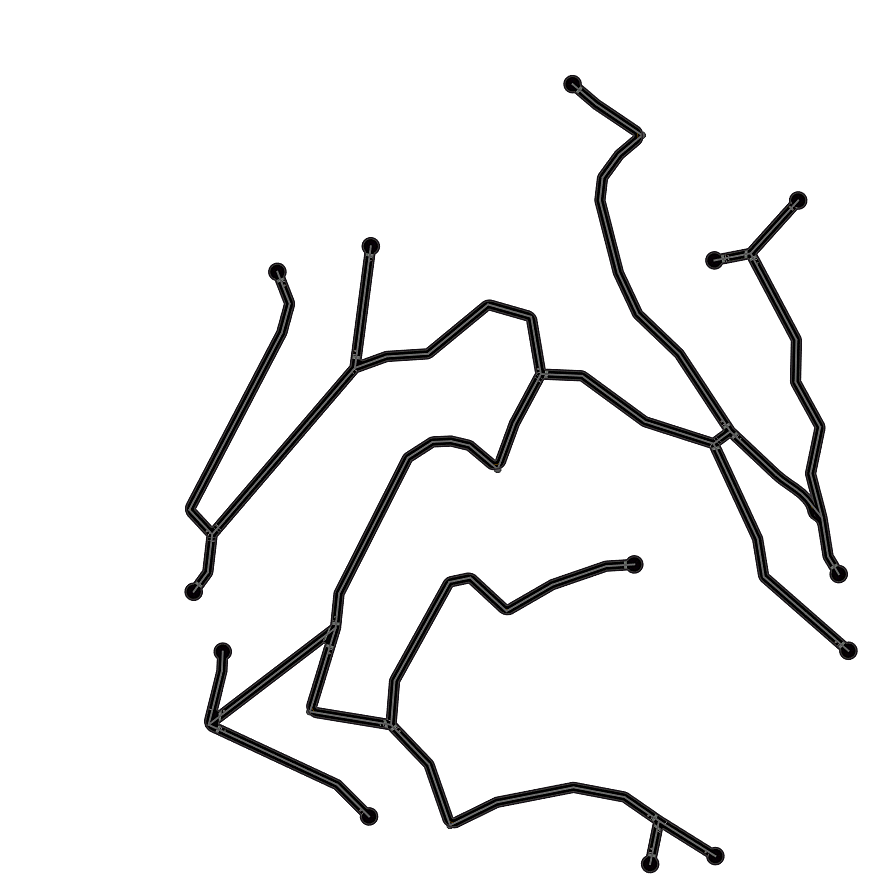}}&
{\includegraphics[width=0.115\linewidth]{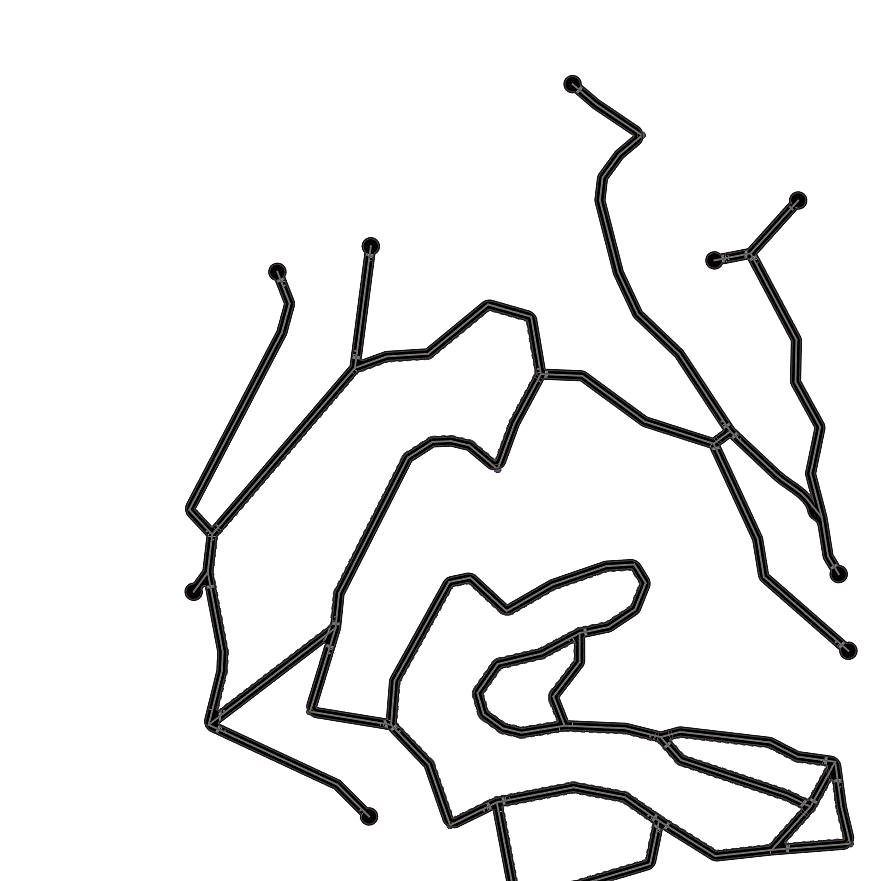}}\\[-0.8mm]
\midrule
{\includegraphics[width=0.115\linewidth]{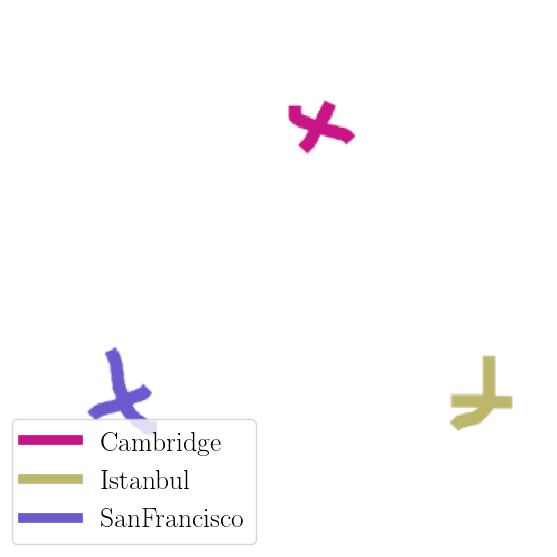}}&
{\includegraphics[width=0.115\linewidth]{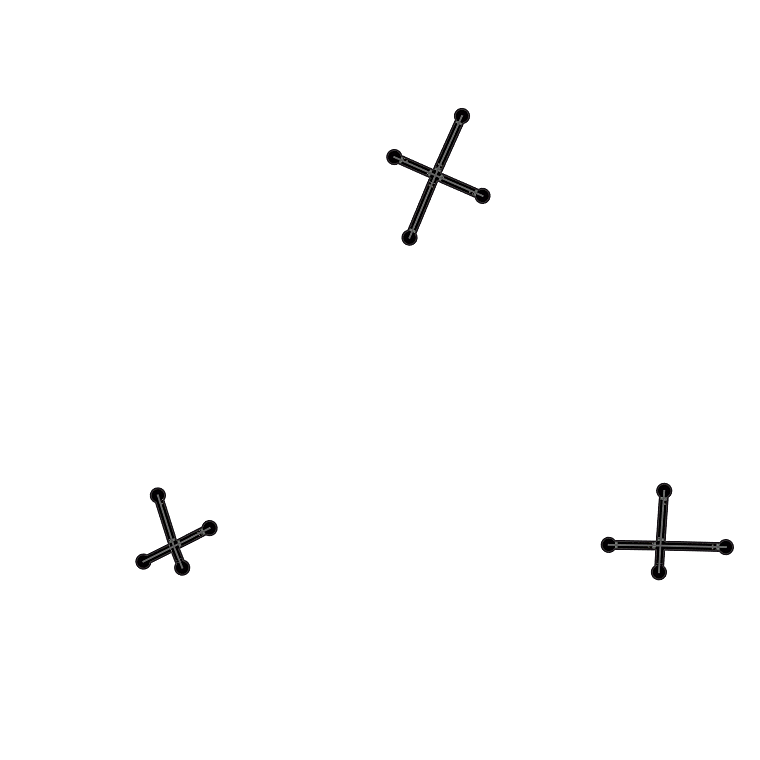}}&
{\includegraphics[width=0.115\linewidth]{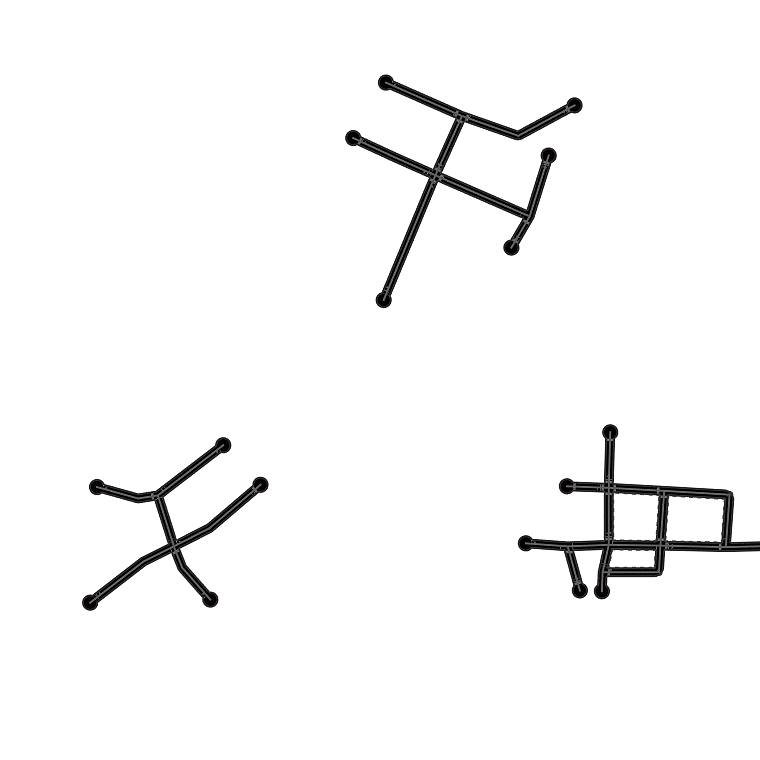}}&
{\includegraphics[width=0.115\linewidth]{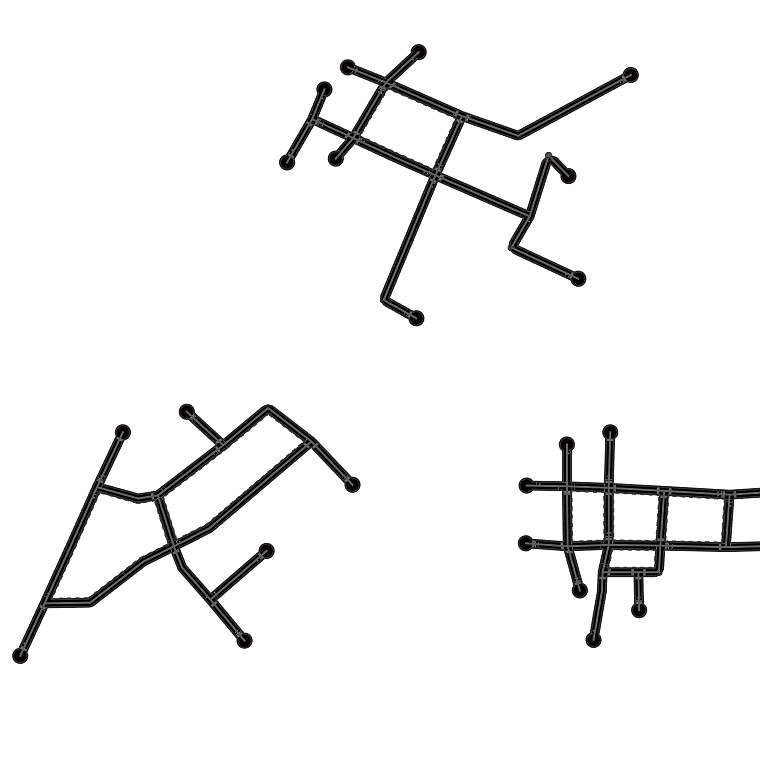}}&
{\includegraphics[width=0.115\linewidth]{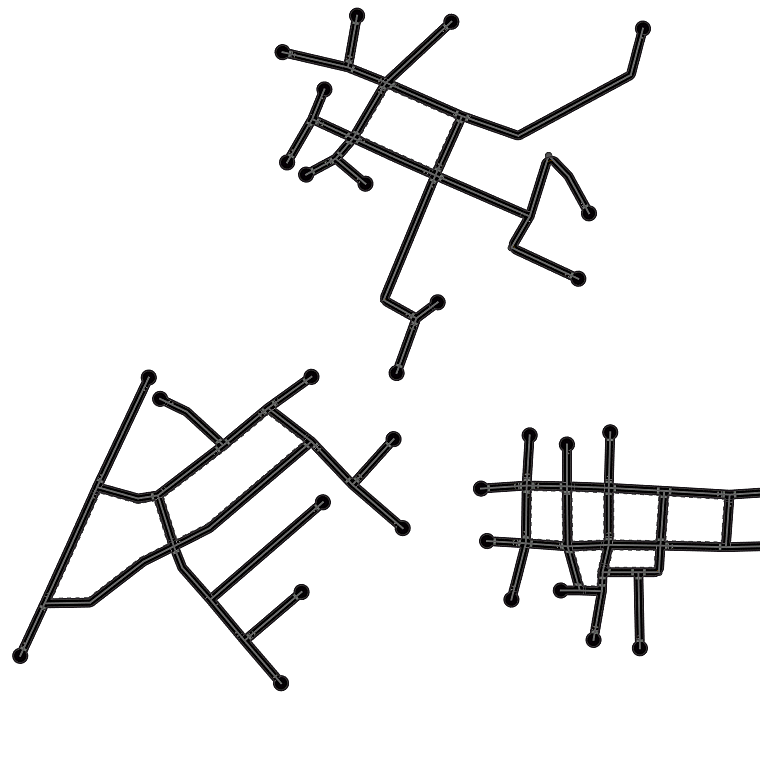}}&
{\includegraphics[width=0.115\linewidth]{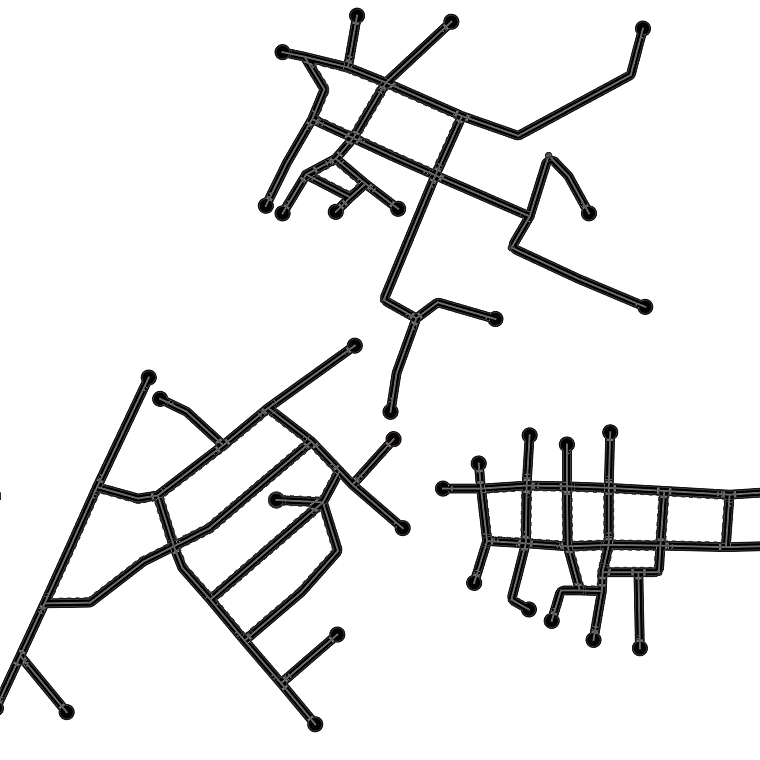}}&
{\includegraphics[width=0.115\linewidth]{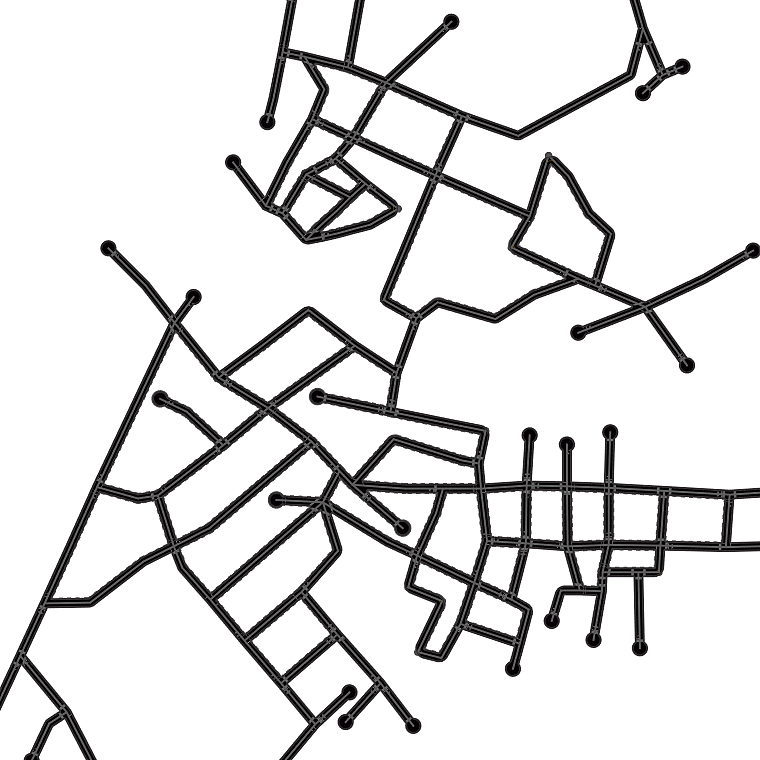}}\\[-0.8mm]
\midrule
\end{tabular}
\end{center}
\vspace{-7mm}
\caption{\small Examples of interactive city road layout generation via user sketching and local style selection.}
\label{fig:interactive}
\vspace{-4mm}
\end{figure*}

\noindent\textbf{Interactive Synthesis.\enspace} We showcase an application for interactive road layout generation where a user chooses from a palette of cities and provides local topology priors by sketching. We match the user's input with pre-stored node templates to form the root node. To allow generating multiple components on the same canvas, we simply modify the NTG inference procedure to iterate through multiple queues in parallel. Fig.~\ref{fig:interactive} shows examples of the generation process. 
\noindent\textbf{Beyond Road Layouts.\enspace} In Appendix, we show results on using NTG's multipath paradigm for learning effective representation of complex shapes, such as multi-stroke hand drawings. This shows potential as a general purpose spatial graph generative model beyond the city road modeling.

%% file: 4_experiment_part2.tex
\subsection{Satellite Road Parsing}
\noindent\textbf{SpaceNet Dataset.\enspace} While several datasets have been presented for road detection~\cite{spacenet,demir2018deepglobe,wang2017torontocity,bastani2018roadtracer},  we use SpaceNet~\cite{spacenet} for its large scale, image quality, and open license. To facilitate consistent future benchmarking, we reformat the raw data into an easy-to-use version with consistent tile size in metric space. Tab.~\ref{tab:stat} shows its statistics. We split tiles of each city into train-validation-test with a 4-1-1 ratio.

\begin{figure*}[h!]
\vspace{-1mm}
\begin{center}
\setlength{\tabcolsep}{2pt}
\begin{tabular}{|c||c|c|c|c|c|c|c|}
\hline
~ & \scriptsize{DRM~\cite{mattyus2017deeproadmapper}} & \scriptsize{RoadExtractor~\cite{bastani2018roadtracer}} & \scriptsize{RoadTracer~\cite{bastani2018roadtracer}} & \scriptsize{FCN~\cite{he2016deep,long2015fully}} & \scriptsize{DLA+STEAL~\cite{yu2018deep,acuna2019steal}} & \scriptsize{NTG} & \tiny{GT}\\
\hline
\midrule[1pt]
\rotatebox{90}{Khartoum} &
{\includegraphics[width=0.115\linewidth]{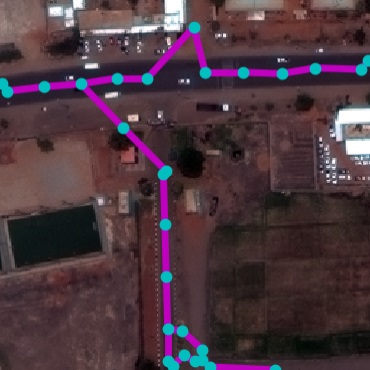}}&
{\includegraphics[width=0.115\linewidth]{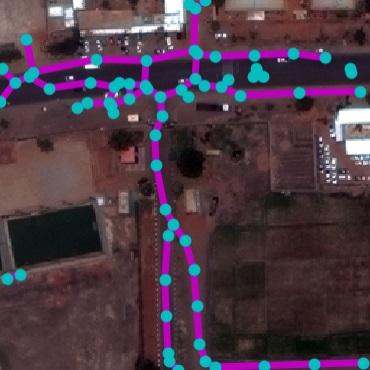}}&
{\includegraphics[width=0.115\linewidth]{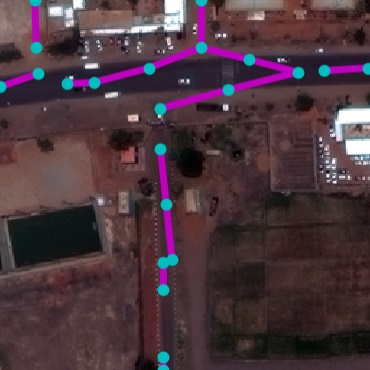}}&
{\includegraphics[width=0.115\linewidth]{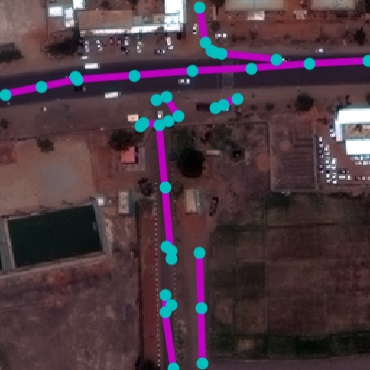}}&
{\includegraphics[width=0.115\linewidth]{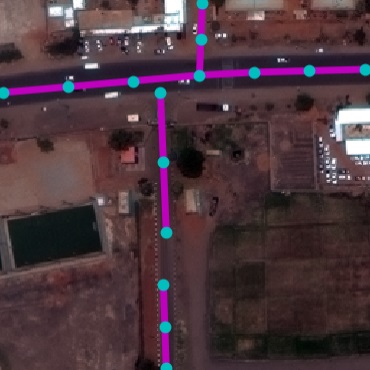}}&
{\includegraphics[width=0.115\linewidth]{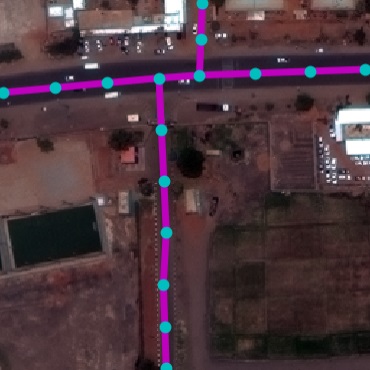}}&
{\includegraphics[width=0.115\linewidth]{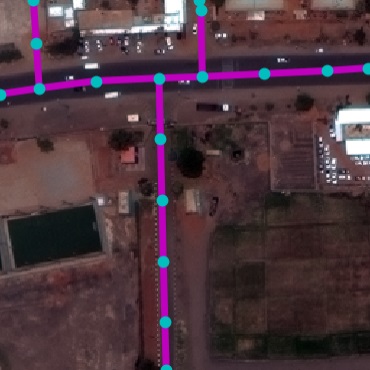}}\\[-0.8mm]
\midrule[1pt]
\rotatebox{90}{Paris} &
{\includegraphics[width=0.115\linewidth]{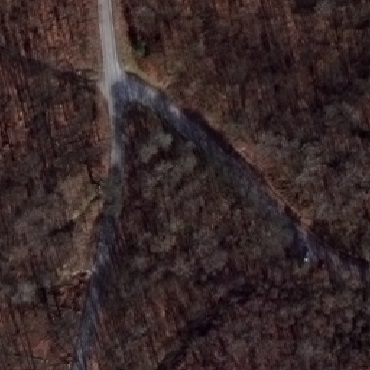}}&
{\includegraphics[width=0.115\linewidth]{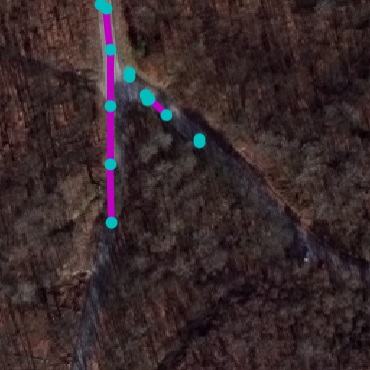}}&
{\includegraphics[width=0.115\linewidth]{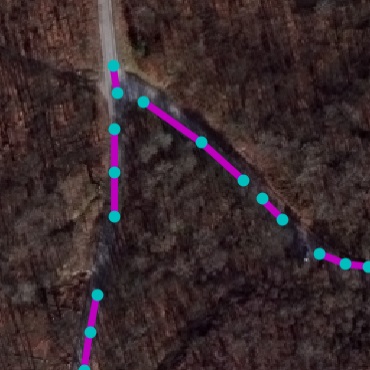}}&
{\includegraphics[width=0.115\linewidth]{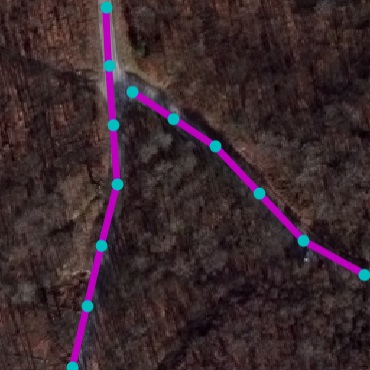}}&
{\includegraphics[width=0.115\linewidth]{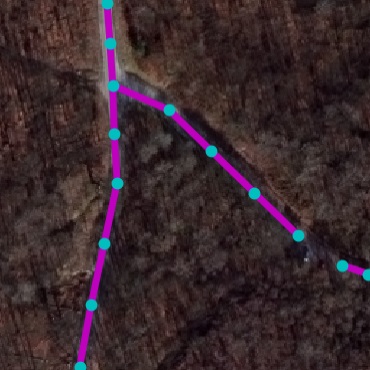}}&
{\includegraphics[width=0.115\linewidth]{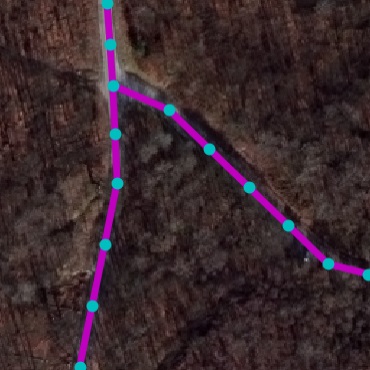}}&
{\includegraphics[width=0.115\linewidth]{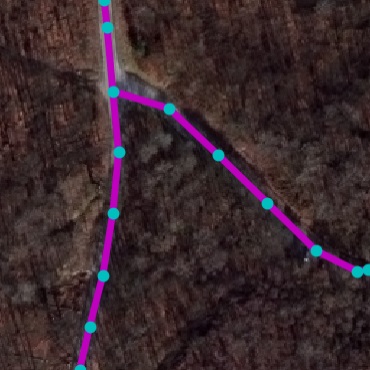}}\\[-0.8mm]
\midrule[1pt]
\rotatebox{90}{Shanghai} &
{\includegraphics[width=0.115\linewidth]{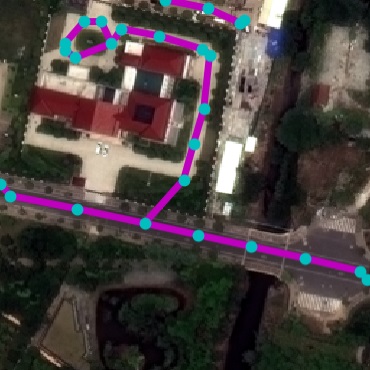}}&
{\includegraphics[width=0.115\linewidth]{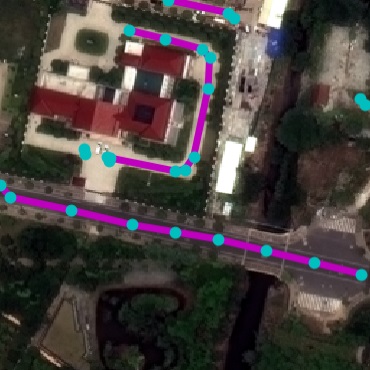}}&
{\includegraphics[width=0.115\linewidth]{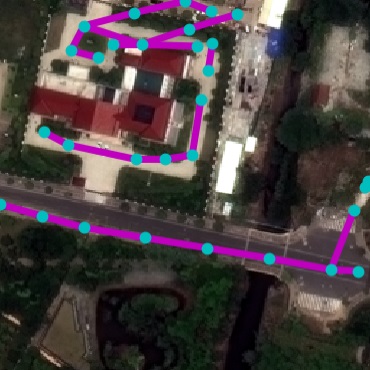}}&
{\includegraphics[width=0.115\linewidth]{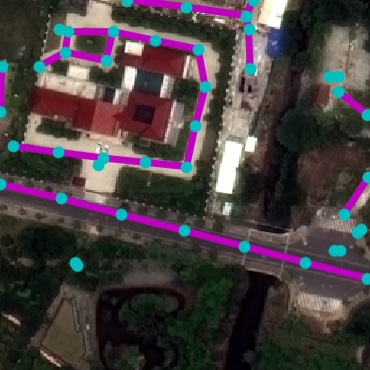}}&
{\includegraphics[width=0.115\linewidth]{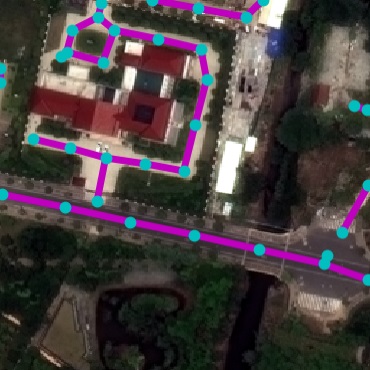}}&
{\includegraphics[width=0.115\linewidth]{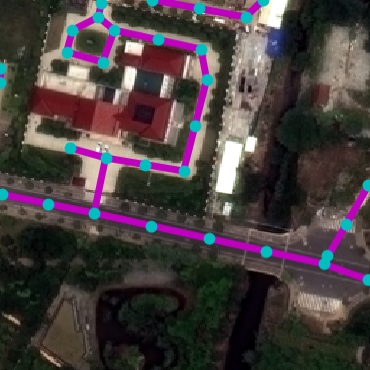}}&
{\includegraphics[width=0.115\linewidth]{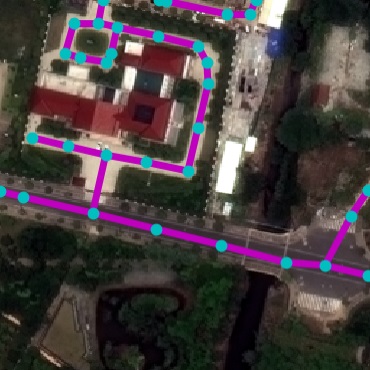}}\\[-0.8mm]
\midrule[1pt]
\rotatebox{90}{Vegas} &
{\includegraphics[width=0.115\linewidth]{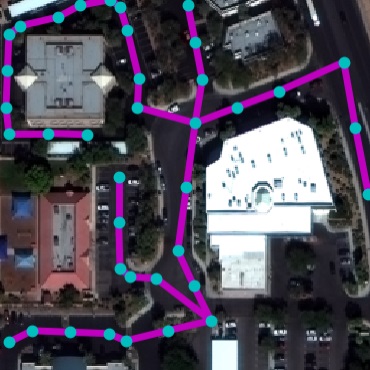}}&
{\includegraphics[width=0.115\linewidth]{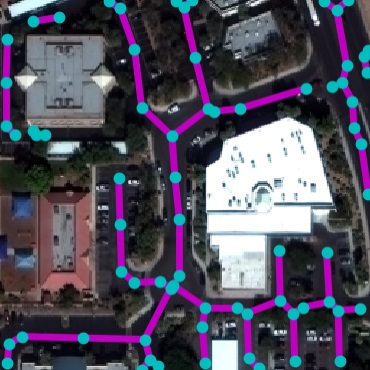}}&
{\includegraphics[width=0.115\linewidth]{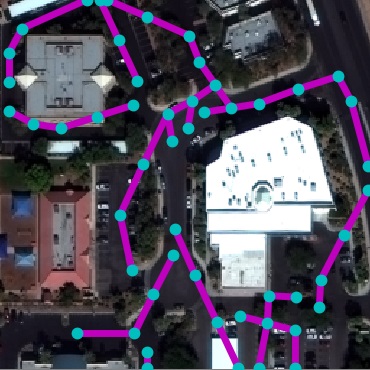}}&
{\includegraphics[width=0.115\linewidth]{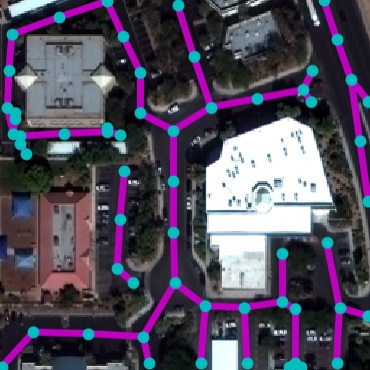}}&
{\includegraphics[width=0.115\linewidth]{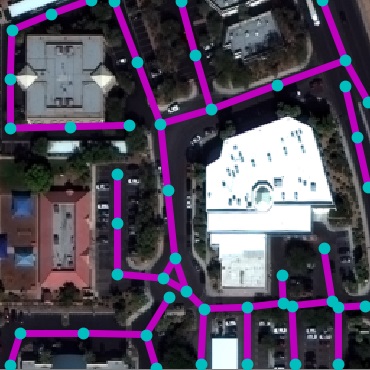}}&
{\includegraphics[width=0.115\linewidth]{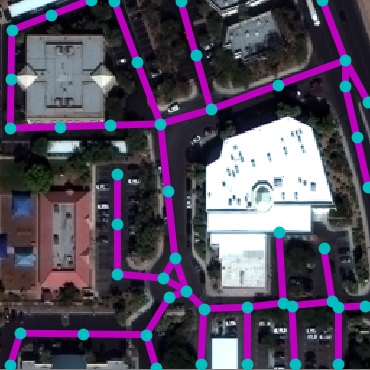}}&
{\includegraphics[width=0.115\linewidth]{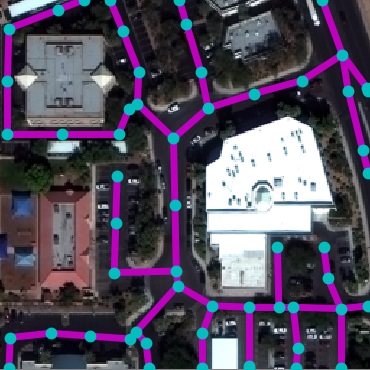}}\\[-0.8mm]
\midrule[1pt]
\end{tabular}
\end{center}
\vspace{-7mm}
\caption{\small Qualitative examples of SpaceNet road parsing.}
\label{fig:roadexample}
\vspace{-3mm}
\end{figure*}

\begin{figure*}[h!]
\begin{center}
\setlength{\tabcolsep}{2pt}
\begin{tabular}{|c||c||c||c|c|c|c|}
\hline
~ & \scriptsize{satellite} & \scriptsize{detection} & \scriptsize{simulation 1} & \scriptsize{simulation 2} & \scriptsize{simulation 3} & \scriptsize{simulation 4}\\
\hline
\midrule[1pt]
\rotatebox{90}{Beijing} &
{\includegraphics[width=0.15\linewidth]{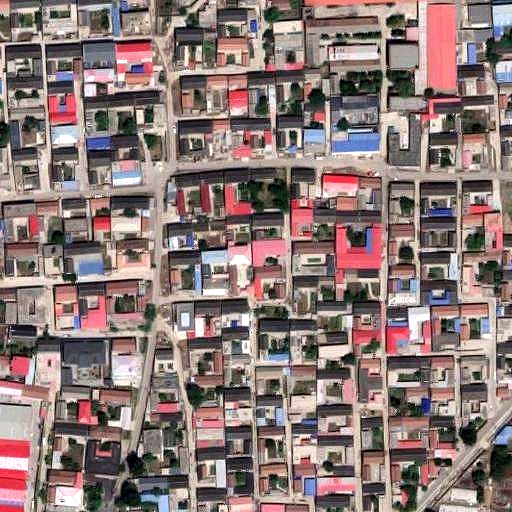}}&
{\includegraphics[width=0.15\linewidth]{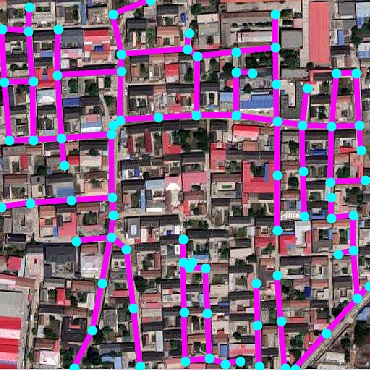}}&
{\includegraphics[width=0.15\linewidth]{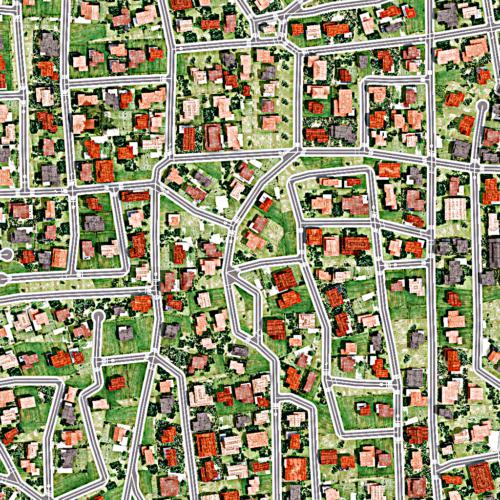}}&
{\includegraphics[width=0.15\linewidth]{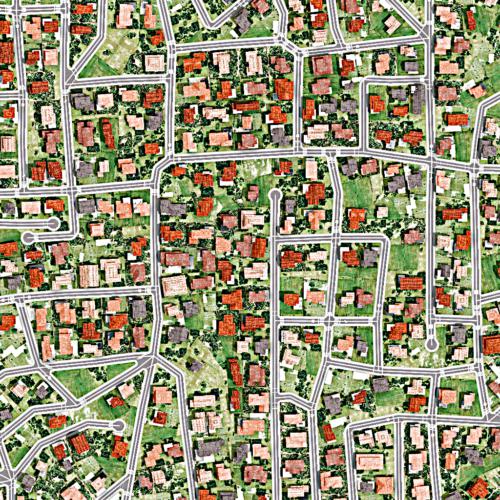}}&
{\includegraphics[width=0.15\linewidth]{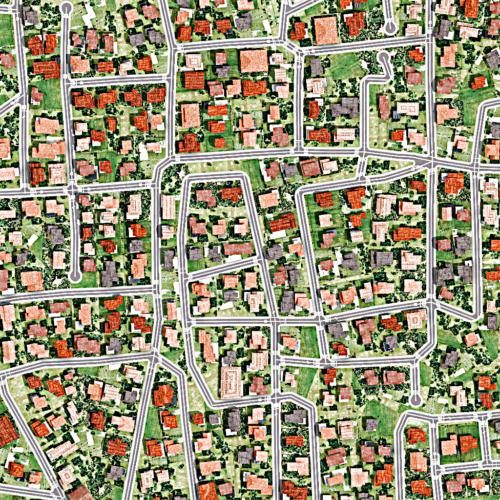}}&
{\includegraphics[width=0.15\linewidth]{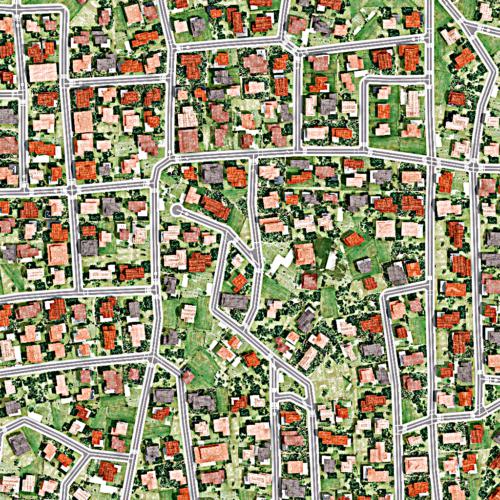}}\\[-0.8mm]
\midrule
\rotatebox{90}{Phoenix} &
{\includegraphics[width=0.15\linewidth]{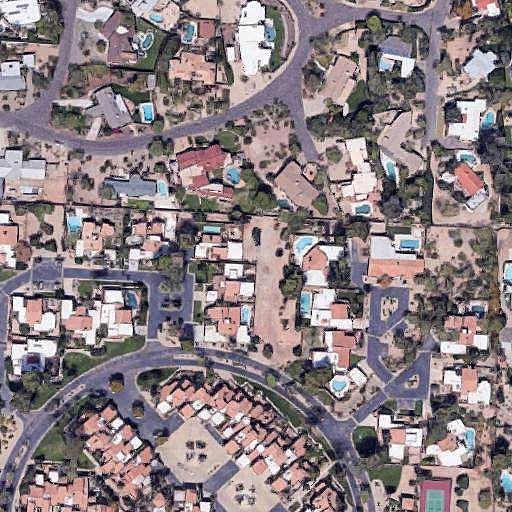}}&
{\includegraphics[width=0.15\linewidth]{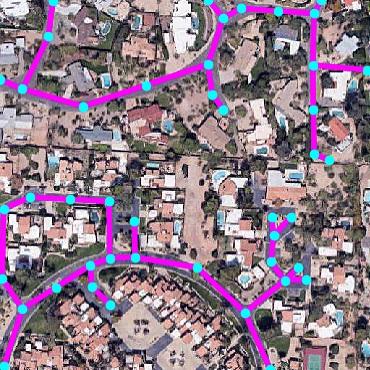}}&
{\includegraphics[width=0.15\linewidth]{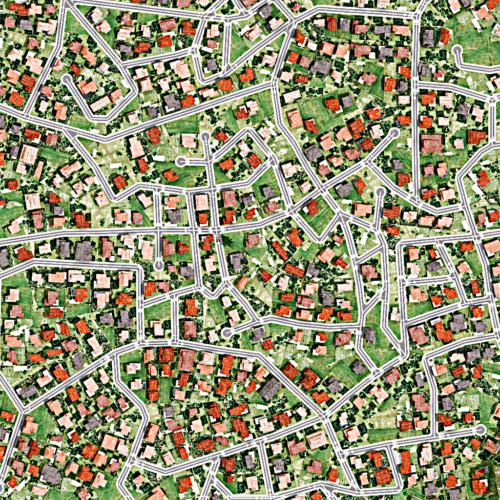}}&
{\includegraphics[width=0.15\linewidth]{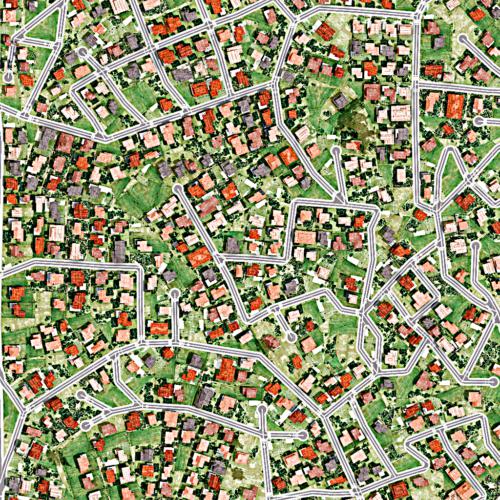}}&
{\includegraphics[width=0.15\linewidth]{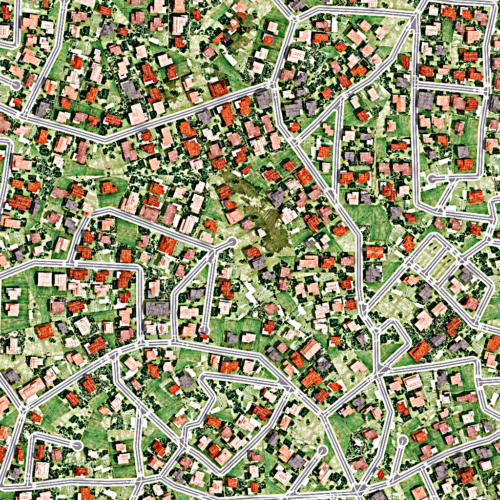}}&
{\includegraphics[width=0.15\linewidth]{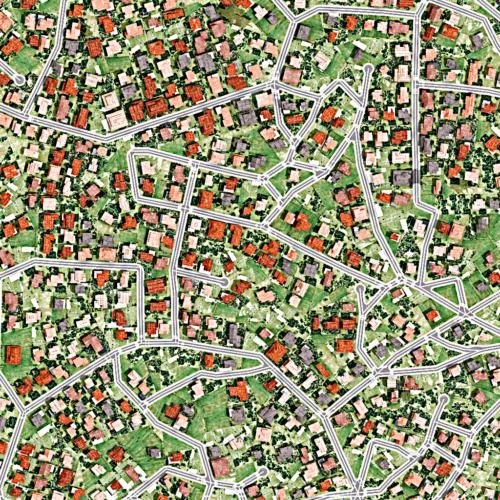}}\\[-0.8mm]
\midrule
\end{tabular}
\end{center}
\vspace{-7mm}
\caption{\small Sat2Sim: converting satellite image into a series of simulated environments. Buildings and vegetation added via~\cite{cityengine}.}
\label{fig:envsim}
\vspace{-4mm}
\end{figure*}

\vspace{-4mm}
\subsubsection{Metrics} 
\vspace{-2mm}
Average Path Length Similarity (APLS) has been shown to be the best metric to reflect routing properties~\cite{spacenet}. Between two graphs, APLS is defined as
\vspace{-3mm}
\begin{equation*}
\vspace{-3mm}
\mathrm{APLS}=1-\frac{1}{N_{p}}\sum_{p_{\mb{v}_1\mb{v}_2}<\infty}min\left\{1,\frac{|p_{\mb{v}_1\mb{v}_2}-p_{\mb{v}_1^{\prime}\mb{v}_2^{\prime}}|}{p_{\mb{v}_1\mb{v}_2}}\right\}
\end{equation*}
where $\mb{v}$ denotes a source graph node, $\mb{v}^{\prime}$ as its closest on-road point in the target graph if such a point exists within a buffer range (5\textit{m}), $N_p$ number of paths. Here, $p_{\mb{v}_1\mb{v}_2}$ denotes the Dijkstra shortest path length between two nodes, and has infinite value if no path exists. We also exchange source and target graphs to establish metric symmetry. To ensure even node distribution, graphs are RDP-simplified~\cite{ramer1972iterative,douglas1973algorithms} and uniformly subdivided with 30\textit{m} maximum edge length. While we use APLS as our main metric, we also report conventional pixel-wise IOU and F1 score as references, even though they are less desirable as revealed in~\cite{spacenet}. 

\vspace{-4mm}
\subsubsection{Results}
\vspace{-2mm}
\ We compare three categories of methods:\\[1mm]
\noindent$\bullet$ \textbf{Prior art}: We evaluate DeepRoadMapper~\cite{mattyus2017deeproadmapper}, RoadExtractor~\cite{bastani2018roadtracer}, and RoadTracer~\cite{bastani2018roadtracer}. RoadTracer requires additional starting points as input. We use the most likely pixel predicted by their CNN, as well as 30 points randomly selected from ground truth (RoadTracer-30).

\noindent$\bullet$ \textbf{Stronger CNNs}: We explore more powerful CNN architectures. We train an FCN with a ResNet backbone~\cite{he2016deep,long2015fully}, as well as a CNN using DLA~\cite{yu2018deep} with STEAL~\cite{acuna2019steal}. To obtain the graph we use standard thinning and geodesic sorting.

\noindent$\bullet$ \textbf{NTG}: 
We evaluate both the parsing NTG (NTG-P) that is only trained on RoadNet and acts as a topological prior and image-based NTG (NTG-I) that is trained on SpaceNet.

\begin{table}[t!]
\vspace{-2mm}
\centering
\small
\begin{tabular}{l||ll|l}
	\hline
	~ & IOU & F1 & APLS\\
	\hline
	\hline
	DeepRoadMapper~\cite{mattyus2017deeproadmapper} & 45.02 & 62.08 & 51.49\\
	RoadExtractor~\cite{bastani2018roadtracer} & 52.91 & 69.20 & 57.38\\
	RoadTracer~\cite{bastani2018roadtracer} & 10.23 & 18.56 & 48.55\\
	RoadTracer-30~\cite{bastani2018roadtracer} & 48.29 & 65.13 & 42.94\\
	FCN~\cite{he2016deep,long2015fully} & 51.09 & 67.63 & 56.56\\
	DLA+STEAL~\cite{yu2018deep,acuna2019steal} & 58.96 & 74.18 & 71.04\\
	\hline
	NTG-P (\cite{mattyus2017deeproadmapper}'s CNN) & 50.58 & 67.18 & 55.87\\
	NTG-P (\cite{bastani2018roadtracer}'s CNN) & 51.62 & 68.09 & 58.79\\
	NTG-P (DLA+STEAL) & 59.29 & 74.44 & 70.99\\
	NTG-I (DLA+STEAL) & \textbf{63.15} & \textbf{77.42} & \textbf{74.97}\\
	\hline
\end{tabular}
\vspace{-3mm}
\caption{\small Comparison of methods on the standard SpaceNet split.}
\label{tab:spaceresult1}

\begin{tabular}{l||ll|l}
	\hline
	~ & IOU & F1 & APLS\\
	\hline
	\hline
	RoadExtractor~\cite{bastani2018roadtracer} & 20.51 & 34.03 & 43.06\\
	DLA+STEAL~\cite{yu2018deep,acuna2019steal} & 33.94 & 50.68 & 56.15\\
	\hline
	NTG-P (DLA+STEAL)& \textbf{35.16} & \textbf{52.02} & \textbf{57.89}\\
	\hline
\end{tabular}
\vspace{-3mm}
\caption{\small SpaceNet evaluation on unseen city by holding one city out in training. Without finetuning, the RoadNet pretrained NTG-P is able to improve over DLA+STEAL.}
\label{tab:spaceresult2}
\vspace{-5mm}
\end{table}

Table~\ref{tab:spaceresult1} and Figure~\ref{fig:roadexample} present SpaceNet results. It can be seen that our method outperforms baselines in all metrics. The DLA+STEAL CNN produces cleaner predictions that focus on road. NTG-P trained only with RoadNet is able to successfully parse graph structure. Using NTG-I that further takes CNN output as input achieves the best result. We also experiment the RoadNet trained NTG-P with CNNs from prior art~\cite{mattyus2017deeproadmapper,bastani2018roadtracer}. It can be seen that the city topology knowledge of NTG makes it a better graph extractor compared to hand-crafted postprocessings in \cite{mattyus2017deeproadmapper,bastani2018roadtracer}, especially in terms of APLS. For NTG-P with DLA+STEAL we notice it has similar performance as standard graph extraction. This is because DLA+STEAL prediction has high confidence as it is trained and tested with same cities that have similar visual appearance. We therefore further experiment with one city held-out to simulate road parsing in unseen cities. Results are presented in Table~\ref{tab:spaceresult2}. It can be seen that NTG-P is able to further improve the result, demonstrating the effectiveness of generative road layout knowledge learnt from RoadNet. We conduct 4-fold evaluation holding out each city per fold, and report the average result.

\vspace{-2mm}
\subsection{Environment Simulation}
\vspace{-2mm}
We further showcase a novel application that combines our two tasks in Figure~\ref{fig:envsim}. 
We propose to directly convert a satellite image into simulation-ready environments, which may be important for testing autonomous vehicles in the future. First, we detect roads in the satellite image with NTG, giving us an initial graph.
Then, we exploit our generative model to propose plausible variations. This is done by pushing all single-connection nodes in the parsed graph into our generative queue, and running the NTG generative process to expand the graph. We directly make use of the NTG model trained for city generation and choose a random city id for each run.
This has two main advantages. First, it is fully automatic and only requires a low-cost satellite image as input. Second, it provides a set of plausible variations of the environment (city) instead of a static one, which could eventually enable training more robust agents. For visualization, we additionally add buildings and tree via~\cite{cityengine}, showing plausible and diverse simulation-ready cities. 

%% file: 5_concolusion.tex
\vspace{-3mm}
\section{Conclusion}
\vspace{-3mm}
In this paper, we proposed Neural Turtle Graphics for generating large spatial graphs. NTG takes the form of an encoder-decoder neural network which operates on graphs locally. 
We showcased NTG on generating plausible new versions of cities, interactive generation of city road layouts, as well as aerial road parsing. Furthermore, we combined the two tasks of aerial parsing and generation, and highlighted NTG to automatically simulate  new cities for which it has not seen any part of the map during training time. In future work, we aim to tackle generation of other city elements such as buildings and vegetation.